\documentclass[final,onefignum,onetabnum]{arxiv}

\usepackage{lipsum}
\usepackage{amsfonts}
\usepackage{graphicx}
\usepackage{epstopdf}
\usepackage{algorithmic}

\usepackage{amsmath,amssymb}
\usepackage{booktabs}
\usepackage{enumerate}

\newcommand{\less}{\leqslant}
\newcommand{\gre}{\geqslant}

\newcommand{\wtilde}{\widetilde}
\newcommand{\argmin}{\mathop{\rm argmin}}

\newcommand{\defn}{\ensuremath{: \, =}}
\newcommand{\real}{\ensuremath{\mathbb{R}}}
\newcommand{\xstar}{\ensuremath{\mathbf{x^*}}}

\newsiamremark{remark}{Remark}
\newsiamremark{hypothesis}{Hypothesis}
\newsiamremark{assumption}{Assumption}
\newsiamremark{condition}{Condition}
\crefname{hypothesis}{Hypothesis}{Hypotheses}
\newsiamthm{claim}{Claim}

\headers{Adaptive Sketching-based Solvers}{J. Lacotte, and M. Pilanci}

\title{Fast Convex Quadratic Optimization Solvers with Adaptive Sketching-based Preconditioners}

\author{Jonathan Lacotte\thanks{Department of Electrical Engineering, Stanford University, 
		(\email{lacotte@stanford.edu}).}
	\and Mert Pilanci\thanks{Department of Electrical Engineering, Stanford University
		(\email{pilanci@stanford.edu}).}}

\usepackage{amsopn}

\ifpdf
\hypersetup{
	pdftitle={Fast Convex Quadratic Optimization Solvers with Adaptive Sketching-based Preconditioners},
	pdfauthor={J. Lacotte, and M. Pilanci}
}
\fi

\begin{document}

\maketitle

\begin{abstract}
We consider least-squares problems with quadratic regularization and propose novel randomized iterative solvers with a sketching-based preconditioner and an adaptive embedding dimension. Without regularization, iterative methods are guaranteed to converge linearly provided that the sketch size scales in terms of the rank. In contrast, with regularization, the sketch size must scale in terms of the effective dimension of the data matrix to guarantee linear convergence. This yields computational savings as the effective dimension is always smaller than the rank, and eventually much smaller for data matrices with fast spectral decay. However, a major difficulty in choosing the sketch size in terms of the effective dimension lies in the fact that the latter is usually unknown in practice. Current sketching-based solvers for regularized least-squares fall short on addressing this issue. Our main contribution is to propose adaptive versions of standard sketching-based iterative solvers, namely, the iterative Hessian sketch and the preconditioned conjugate gradient method, that do not require a priori estimation of the effective dimension. We propose an adaptive mechanism to control the sketch size according to the progress made in each step of the iterative solver. If enough progress is not made, the sketch size increases to improve the convergence rate. We prove that the adaptive sketch size scales at most in terms of the effective dimension, and that our adaptive methods are guaranteed to converge linearly. Consequently, our adaptive methods improve the state-of-the-art complexity for solving dense, ill-conditioned least-squares problems. Importantly, we illustrate numerically on several synthetic and real datasets that our method is extremely efficient and is often significantly faster than standard least-squares solvers such as a direct factorization based solver, the conjugate gradient method and its preconditioned variants.
\end{abstract}


\section{Introduction}
\label{SectionIntroduction}

We consider\footnote{This work is based on the preliminary results~\cite{lacotte2020effective} published in the Proceedings of the 34th Conference on Neural Information Processing Systems (NeurIPS 2020).} the convex quadratic program
\begin{align}
\label{eqnmainoptimizationproblem}
    x^* \defn \argmin_{x \in \real^d} \big\{f(x) \defn \frac{1}{2} \langle x, H x\rangle - b^\top x\big\}\,,
\end{align}
where $H = A^\top A + \nu^2 \Lambda$ for a given data matrix $A \in \real^{n \times d}$ with $n \gg 1$, a given vector $b \in \real^d$, a diagonal matrix $\Lambda \succeq I_d$ and a parameter $ \nu > 0$. By considering instead the dual problem of~\eqref{eqnmainoptimizationproblem},
\begin{align}
    \label{eqndualoptimizationproblem}    
    w^* \defn \min_{z \in \real^n} \big\{\frac{1}{2}\langle w, ((A\Lambda^{-\frac{1}{2}}) (A\Lambda^{-\frac{1}{2}})^\top + \nu^2 \, I_n) w\rangle - (A\Lambda^{-1} b)^\top w\big\}\,,
\end{align}
we can assume without loss of generality that $n \gre d$. Typical applications of formulation~\eqref{eqnmainoptimizationproblem} are ridge regression and classical instances of Newton linear systems. In this work, we are interested in the following class of preconditioned first-order methods
\begin{align}
\label{eqnfirstordermethods}
    x_{t+1} \in x_0 + H_S^{-1} \cdot \textrm{span}\!\left\{\nabla f(x_0), \hdots, \nabla f(x_t)\right\}\,,
\end{align}
where $x_0 \in \real^d$, the matrix $S$ is $m \times n$ with embedding dimension $m \ll n$ and the approximate Hessian $H_S$ is defined as $H_S \defn A^\top S^\top S A + \nu^2 \cdot \Lambda$. We consider a \emph{random embedding} (or \emph{sketching matrix}) $S$. Classical random embeddings include Gaussian embeddings with independent entries $\mathcal{N}(0,1/m)$, the subsampled randomized Hadamard transform (SRHT) and the sparse Johnson Lindenstrauss transform (SJLT); see \cref{sectionsketchingmatrices} for background.

Modern randomized iterative methods outperform classical algorithms in solving dense ill-conditioned linear systems \cite{rokhlin2008fast,avron2010blendenpik,pilanci2016iterative}. In the large-scale setting, direct factorization methods have prohibitive computational cost $\mathcal{O}(nd^2)$ whereas the performance of standard iterative solvers~\cite{nocedal2006numerical} such as the conjugate gradient method (CG)~\cite{hestenes1952methods} or Chebyshev iterations~\cite{manteuffel1977tchebychev} have condition number dependency. In this regard, preconditioned first-order methods~\eqref{eqnfirstordermethods}  involve using a random embedding $S \in \real^{m \times n}$ with $m \ll n$ to project the matrix $A$ and improve the condition number. A canonical instance is the iterative Hessian sketch (IHS) (introduced by~\cite{pilanci2016iterative} and considered later in~\cite{ozaslan2019iterative, lacotte2019faster, lacotte2020optimal, ozaslan2019regularized, pilanci2017newton, lacotte2020effective}) whose update is given by
\begin{align}
\label{eqnihs}
    x_{t+1} = x_{t} - \mu_t \cdot H_S^{-1} \nabla f(x_t)\,.
\end{align}
The IHS can be viewed as gradient descent with preconditioner $H_S$. The preconditioned conjugate gradient method (PCG) (see, e.g.,~\cite{nocedal2006numerical} for background) is also an instance of~\eqref{eqnfirstordermethods}. Specialized to the preconditioner $H_S$, PCG starts with $r_0 = b - H x_0$, solves  $H_S \wtilde r_0 = r_0$, and sets $p_0 = \wtilde r_0$ and $\wtilde \delta_0 = r_0^\top \wtilde r_0$. Then, it computes at each iteration
\begin{align}
\label{eqnpcgupdate}
    x_{t+1} = x_t + \alpha_t p_t\,,\quad r_{t+1} = r_t - \alpha_t H p_t\,,\quad p_{t+1} = \wtilde r_{t+1} + (\wtilde \delta_{t+1}/\wtilde \delta_t) \, p_t\,,
\end{align}
where $\alpha_t = \wtilde \delta_t / p_t^\top H p_t$, $\wtilde \delta_{t+1} = r_{t+1}^\top \wtilde r_{t+1}$, and $\wtilde r_{t+1}$ is the solution of the linear system $H_S \wtilde r_{t+1} = r_{t+1}$. Efficient implementations of the IHS and PCG are based on precomputing a factorization (e.g., in terms of diagonal, orthogonal and/or triangular matrices) of $H_S$ such that the linear system can be efficiently solved; see \cref{sectioncomplexity} for details.

The choice of the sketch size $m$ is critical for the performance of sketching-based iterative solvers. In the \emph{unregularized} case ($\nu=0$), the sketch size must usually scale in terms of the rank $d$. In fact, the authors of~\cite{ozaslan2019iterative, lacotte2019faster} showed that the IHS accelerated with heavy-ball momentum and with, for instance, Gaussian embeddings satisfies $\|x_t-x^*\|^2_H \lesssim (d/m)^t$ (where $\|z\|_H^2 = z^\top H z$) for $m \gtrsim d$. In the \emph{regularized} case ($\nu > 0$), more relevant than the matrix rank is the \emph{effective dimension} $d_e = \text{trace}(A_\nu) / \|A_\nu\|_2$ where $A_\nu = A^\top A (A^\top A + \nu^2 \Lambda)^{-1}$. It satisfies $d_e \less d$ and is eventually significantly smaller than the rank $d$ for a data matrix $A$ with fast spectral decay. In fact, for regularized least-squares ($\Lambda=I_d$), the authors of~\cite{ozaslan2019regularized} have shown for the IHS with, for instance, Gaussian embeddings that $\|x_t - x^*\|_H^2 \lesssim (d_e/m)^t$ for $m \gtrsim d_e$. Moreover, the approximate Hessian $H_S$ can then be factorized in time $\mathcal{O}(d_e^2 d)$ instead of $\mathcal{O}(d^3)$ (see, e.g., the Woodbury matrix identity~\cite{hager1989updating}). A similar result for undetermined ridge regression has been established in~\cite{chowdhury2018iterative} in terms of the effective dimension. In the related context of kernel ridge regression, it was also shown that Nystrom approximations of kernel matrices have performance guarantees for sketch sizes proportional to the effective dimension~\cite{bach2013sharp, alaoui2015fast, cohen2017input}. Lastly, we mention another sketch-and-solve approach which computes $\widetilde x \defn \argmin_x \frac{1}{2} \|SA x - Sb\|_2^2 + \frac{\lambda}{2} \|x\|_2^2$ (see e.g. \cite{drineas2016randnla,pilanci2015randomized, pilanci2016fast, sridhar2020lower, derezinski2020debiasing, bartan2020distributedaveraging}). In~\cite{avron2017sharper}, the authors showed that for $m \approx d_e / \varepsilon$, the estimate $\widetilde x$ satisfies $f(\widetilde x) \less (1+\varepsilon) f(x^*)$. This can result in large $m$ for even medium accuracy, whereas preconditioned iterative solvers yield an $\varepsilon$-approximate solution with $m \approx d_e$ at the expense of an iterative procedure. \cite{wang2017sketched} addressed overdetermined ridge regression where their main motivation is to deal with ill-conditioned data matrices. Differently from us, their sketching algorithms require the sketch size to scale in terms of the rank instead of the effective dimension (see Table 2 therein). 

The major difficulty in choosing $m$ in terms of $d_e$ lies in the estimation of $d_e$ which is usually unknown in practice, and the recent literature falls short on providing efficient procedures for estimating the effective dimension $d_e$ with theoretical guarantees.~\cite{ozaslan2019regularized} proposed a heuristic estimator without estimation guarantees;~\cite{chowdhury2018iterative} does not explicitly address the problem of estimating $d_e$;~\cite{avron2017sharper} proposed an efficient method with provable guarantees but under the restrictive assumption that $d_e$ is very small, i.e., $d_e \less (n+d)^{\frac{1}{3}}/\mbox{poly}(\log(n+d))$ (see Theorem 60 therein). 

Our main goal in this paper is to design an \emph{adaptive} algorithm which does not require a priori estimation of $d_e$, but is still able to use a sketch size $m$ which scales in terms of $d_e$ and achieve a fast convergence rate. Our adaptive method can detect when the progress made at each iteration is insufficient in which case it increases the sketch size. As soon as the adaptive sketch size $m$ scales appropriately in terms of $d_e$, our method is guaranteed to not increase the sketch size anymore and have fast convergence rate. Adaptivity in the context of randomized numerical linear algebra has been considered for other applications, e.g., in randomized range finding~\cite{halko2011finding}. As the algorithm performance depends on both the embedding dimension and the (usually unknown) spectral decay of the involved matrix, a posteriori error estimates were designed by~\cite{martinsson2020randomized}: if the error estimates are unacceptably large, then the embedding dimension is increased. The algorithm is run again until acceptable error is achieved. Our proposed method draws similarities with this approach: at each iteration, we estimate the progress of a new iterate and adapt the sketch size accordingly.

\subsection{Notations}

We denote the Euclidean norm of a vector $z$ by $\|z\|$ or $\|z\|_2$, and we define the norm induced by a positive definite matrix $P$ as $\|z\|_P = \sqrt{z^\top P z}$. We denote the operator norm of a matrix (i.e., its maximum singular value) $M$ by $\|M\|_2$, its Frobenius norm by $\|M\|_F$ and its trace by $\mbox{tr}(M)$. We also use $\sigma_\text{max}(M)$ (resp.~$\sigma_\text{min}(M)$) for its maximum (resp.~minimum) singular value, and $\kappa(M) = \sigma_\text{max}(M)/\sigma_\text{min}(M)$ for its condition number. For a symmetric matrix $P$, we use $\lambda_\text{max}(P)$ (resp.~$\lambda_\text{min}(P)$) to denote the maximum (resp.~minimum) eigenvalue of $P$. Given an eigenvalue decomposition $P=\sum_{i} \gamma_i w_i w_i^\top$, we denote by $P_+$ the matrix $P=\sum_{i} (\gamma_i)_+ w_i w_i^\top$ where $(\gamma)_+ \defn \max\{\gamma,0\}$. If $P$ is positive semi definite, we denote by $P^\frac{1}{2}$ a positive semi definite square-root. We reserve the thin SVD notation $A (A^\top A + \nu^2 \, \Lambda)^{-1/2} = U D V^\top$ where $D$ is diagonal with non-negative entries and $U \in \real^{n \times d}$ and $V \in \real^{d \times d}$ have orthonormal columns. Note that $d_e \defn \frac{\|D\|_F^2}{\|D\|_2^2}$. 

We introduce the positive definite matrix $C_S \defn H^{-\frac{1}{2}} H_S H^{-\frac{1}{2}}$ which plays a critical role in our analysis. Note that $C_S - I_d = H^{-\frac{1}{2}} (H_S - H) H^{-\frac{1}{2}} = V D (U^\top S^\top S U - I_d) D V^\top$. We reserve the notation $C_S = \sum_{i=1}^d \lambda_i v_i v_i^\top$ for an eigenvalue decomposition of $C_S$ where the eigenvalues $\lambda_1 \gre \dots \gre \lambda_d > 0$ are indexed in non-increasing order.

Given $x\in \real^d$, we define the error at $x$ as $\delta_x \defn \frac{1}{2}\,\|x-\xstar\|_H^2$, and we use the shorthand $\delta_t \equiv \delta_{x_t}$ for an iterate $x_t$ at time $t$. We introduce the error vector $\Delta_t = H^\frac{1}{2}(x_t - x^*)$. Note that $\delta_t = \frac{1}{2}\|\Delta_t\|_2^2$.

Given $\rho \in (0,1)$ and $\alpha \gre 0$, we denote $c(\alpha, \rho) \defn \frac{1+\sqrt{\rho}}{1-\sqrt{\rho}} \cdot \alpha$. Informally, we reserve the notations $m$ for sketch sizes, $\mu$ for step sizes, $\beta$ for momentum parameters, $\varepsilon$ for precisions, $\delta \in (0,1)$ for failure probabilities, $\rho > 0$ for subspace embedding approximations, and $P$ and $Q$ for polynomials. We denote by $\mathbb{R}_t[X]$ the set of real polynomials with degree less than $t$.

\subsection{Overview of our Contributions}
\label{sectioncontributions}

\subsubsection*{Adaptive sketch size iterative solvers} Our main contribution is to propose adaptive versions of the IHS and PCG that do not require a priori knowledge of the effective dimension $d_e$ and still achieve a similar linear convergence rate with a time-varying sketch size $m_t$ which is guaranteed to scale at most in terms of the effective dimension $d_e$ (e.g., $m_t \lesssim d_e \log(d_e)$ for the SRHT). Our prototype method (\cref{AlgorithmAdaptive}) is initialized with a small sketch size $m_0$ (e.g., $m_0=1$) and, at each iteration, it uses an improvement criterion to decide whether to increase $m_t$ by a factor $2$ or not. Consequently, we improve for most instances of~\eqref{eqnmainoptimizationproblem} the time and memory complexities to reach an $\varepsilon$-accurate solution: for example, our adaptive PCG (\cref{algorithmadaptivepcg}) with the SRHT verifies $m_t \lesssim d_e \log d_e$ and it returns an $\varepsilon$-accurate solution with high probability (w.h.p.) in time complexity
\begin{align}
\label{eqntimecomplexitysrhtadaptiveintro}
    \mathcal{O}\big(nd\log(1/\varepsilon) + \log d_e \cdot (nd \log d_e + \min\{d, d_e \log d_e\} d d_e \log d_e)\big)\,.
\end{align}
whereas state-of-the-art sketching-based iterative solvers (PCG with $m \asymp d \log d$) have time complexity
\begin{align}
\label{eqntimecomplexitysrhtnonadaptiveintro}
    \mathcal{O}\!\left(nd\log(1/\varepsilon) + nd\log d + d^3 \log d\right)\,.
\end{align}
When $d_e \ll d$, our method has significantly smaller time complexity. Note that when $d \approx d_e$, the factorization and sketching costs $\mathcal{O}(\min\{d, d_e \log d_e\}d d_e\log d_e)$ and $\mathcal{O}(nd \log d_e)$ in~\eqref{eqntimecomplexitysrhtadaptiveintro} of our method come with an additional factor $\log d_e$ compared to~\eqref{eqntimecomplexitysrhtnonadaptiveintro}. This factor is the cost of adaptivity, i.e., it is the maximum number of times we need to increase the sketch size (and thus form a new sketch $S_t A$ and a new factorization of $H_{S_t}$) before reaching the threshold value $d_e \log(d_e)$ that ensures strong embedding properties w.h.p. We evaluate numerically our adaptive algorithms for ridge regression on several synthetic and real datasets, and we show that our method is able to offer major memory savings and speed-ups compared to their non-adaptive counterparts.

\subsubsection*{Subspace embedding properties and sharp concentration bounds} 
Our adaptive methods depend on subspace embedding properties of the random projection $S$ as measured by the deviations of the matrix $C_S$ around the identity matrix $I_d$. It is then critical for optimal design and predictability of our algorithms to quantify precisely these deviations. For Gaussian embeddings, a classical bound~\cite{koltchinskii2017concentration} is given as $\|C_S - I_d\|_2 =\mathcal{O}\big(\|D\|_2^2 \, \sqrt{\rho} \big)$ for $m = \mathcal{O}(d_e/\rho)$ and with probability $1-e^{-\mathcal{O}(d_e)}$. We provide a similar bound with an explicit and sharp characterization of the numerical constants involved. Our analysis is based on a recent extension~\cite{thrampoulidis2014tight} of Gordon's min-max theorem~\cite{gordon1985some}. For the SRHT, the classical results and analysis proposed in~\cite{tropp2011improved} state that $\|U^\top S^\top S U - I_d\|_2 \less \mathcal{O}(\sqrt{\rho})$ for $m=\mathcal{O}(d \log(d)/\rho)$ and with probability $1-\mathcal{O}(1/d)$. We generalize this result and analysis to the effective dimension case, and we explicit the numerical constants involved. We emphasize that the results of~\cite{tropp2011improved} have already been improved and generalized by~\cite{cohen2016optimal} to the effective dimension case (see Theorems~1 and~9) where it is shown that $\|C_S - I_d\|_2 = \mathcal{O}(\|D\|_2^2 \cdot \sqrt{\rho})$ for $m = \mathcal{O}((d_e + \log^2(d_e))/\rho)$ and with probability $1-\mathcal{O}(1/d_e)$. However, their results do not explicit the numerical constants involved and their technical approach is different.

\subsubsection*{Comparison to a preliminary version of this work}

Our present contributions significantly extend the results of a preliminary version \cite{lacotte2020effective} of this work. In particular, we propose an adaptive version of PCG (\cref{algorithmadaptivepcg}) whereas in \cite{lacotte2020effective} we only considered the IHS; we provide a more general and comprehensive framework for adaptivity, that is, our main result (\cref{theoremadaptive}) is stated in terms of any preconditioned first order method that satisfies a linear convergence guarantee (see \cref{conditionlinearconvergence}); we investigate theoretically an adaptive version of the IHS with heavy-ball momentum (\cref{corollaryfinitetimeguaranteespolyak}) whereas in \cite{lacotte2020effective} we solely proposed a numerical investigation; we provide much more extensive numerical comparisons of our adaptive methods with an additional embedding matrix (the SJLT), baseline (a direct factorization method) and datasets. Additionally, we have generalized and also significantly simplified and polished the analysis of \cref{theoremadaptive}, \cref{theoremconcentrationshrt} and \cref{theoremgaussianconcentration} presented in \cite{lacotte2020effective}.

\subsection{Open Questions}

Our methods are guaranteed to converge to an accurate solution with high probability and with improved memory and time complexities compared to state-of-the-art sketching-based solvers. In fact, the sketch size may remain significantly smaller than the effective dimension (see numerical experiments in \cref{sectionnumericalexperiments}), whence the adaptive sketch size does not provide an estimate of the effective dimension. It is left open to understand more precisely how the adaptive sketch size evolves throughout the algorithm, and whether one can estimate the effective dimension from the final sketch size. 

We do not prescribe a termination criterion for our adaptive methods to reach an $\varepsilon$-accurate solution with high-probability. The main difficulty to design one lies in the fact that the sketch size may remain much smaller than the effective dimension. It is in fact straightforward to design an (overly) conservative termination criterion (see \cref{remarkterminationcriterion}), and we leave open the design of a finer criterion. Alternatively, one can simply use as a termination criterion the objective value $f(x_t)$, but this leaves a theoretical gap with the measure of error $\delta_t = \frac{1}{2}\|x_t-x^*\|_H^2$ we use in our theoretical guarantees because $\delta_t \neq f(x_t)$ except for consistent linear systems.

\section{Preliminaries}
\label{sectionpreliminaries}

\subsection{Randomized Sketches}
\label{sectionsketchingmatrices}

Gaussian embeddings, i.e., matrices $S \in \real^{m \times n}$ with i.i.d.~Gaussian entries $\mathcal{N}(0,1/m)$ are a classical random projection whose spectral and subspace embedding properties are tightly characterized. The cost of forming the sketch $S \cdot A$ for a dense matrix $A$ requires $\mathcal{O}(ndm)$ flops (using classical matrix multiplication). In practice, e.g., parallelized computation or for a sparse matrix $A$, the running time can actually be significantly faster (see, e.g.,~\cite{meng2014lsrn} for a detailed discussion of practical advantages of Gaussian embeddings). Sub-Gaussian embeddings are also common (e.g., independent Rademacher entries), and have similar sketching cost and subspace embedding properties, although less finely characterized. On the other hand, a classical orthogonal transform is the SRHT~\cite{ailon2006approximate}, and it provides in general a more favorable trade-off in terms of subspace embedding properties and sketching time. A matrix $S \in \real^{m \times n}$ is a SRHT if $S = R H E$ where $R \in \real^{m \times n}$ is a row-subsampling matrix (uniformly at random without replacement), $H \in \real^{n \times n}$ is the Hadamard matrix\footnote{The Hadamard transform is defined for $n = 2^k$ for some $k \gre 0$. If $n$ is not a power of $2$, a standard practice is to pad the data matrix with $2^{\lceil \log_2(n)\rceil}-n$ additional rows of $0$'s.} of size $n$, and $E$ is a diagonal matrix with random signs on the diagonal. Its sketching cost is near-linear and scales as $\mathcal{O}(nd\log m)$. Another common choice in practice is a sparse embedding which allows for even faster sketching. Specifically, we consider the SJLT~\cite{clarkson2017low}: for each column, a number $s$ of rows are chosen uniformly at random without replacement, and the corresponding entries are randomly chosen in $\{\pm 1/\sqrt{s}\}$. The sketching cost scales as $\mathcal{O}(s \cdot \text{nnz}(A))$ (where $\text{nnz}(A)$ is the number of non-zero entries of $A$) independently of the sketch size $m$. However, to ensure strong subspace embedding properties, there is a trade-off between $s$ and the sketch size $m$. For instance, the choice $s\gtrsim \log(d/\sqrt{\rho})$ and $m \gtrsim d \log(d) / \rho$ ensures that $\|U^\top S^\top SU-I_d\|_2 \less \sqrt{\rho}$, whereas $s=1$ requires $m \gtrsim d^2 / \rho$. We choose the SJLT with $s=1$ for its simplicity, but our analysis readily extends to any $s \gre 1$ and to the more general class of OSNAPs~\cite{nelson2013osnap}.

\subsection{Subspace Embedding Properties}

Given an embedding matrix $S \in \real^{m \times d}$ and a deviation parameter $\rho > 0$, we define the $S$-measurable event 
\begin{align}
    \mathcal{E}_{\rho}^m \defn \Big\{ \|C_S - I_d\|_2 \less \max\{\sqrt{\rho}, \rho\} \Big\}\,.
\end{align}
Given an additional parameter $\delta \in (0,1)$, we define the critical sketch size 
\begin{align}
    m_\delta \defn \inf \left\{ k \gre 1 \mid \mathbb{P}(\mathcal{E}_{\rho}^m) \gre 1-\delta\,,\,\, \forall \,m \gre k/\rho,\,\forall \rho > 0 \right\}\,.
\end{align}
In \cref{tableembeddingproperties}, we provide upper bounds on $m_\delta$ for the random embeddings we consider. These results are standard: we refer to, for instance, the analysis of~\cite{cohen2016optimal} based on the so-called oblivious subspace embedding moment property which extends embedding properties over Euclidean balls to ellipsoids; see also~\cite{nelson2013osnap} for the more general sparse embeddings~OSNAPs.  We provide additional efforts to sharpen the constants involved in some of those bounds, and we state precisely these results later in \cref{sectionconcentration}. For the SRHT, our technical innovation is a novel variation of classical matrix Chernoff bounds~\cite{tropp2011improved, tropp2015introduction}. For Gaussian embeddings which can be more finely analyzed than sub-Gaussian ones, we leverage the convex Gordon min-max theorem~\cite{gordon1985some, thrampoulidis2014tight}.
\begin{table}[!h]
\caption{Critical sketch size $m_{\delta}$ for some classical random embeddings: the SRHT, the SJLT with $s=1$ non-zero entry per column sampled uniformly at random without replacement and sub-Gaussian embeddings.}
\label{tableembeddingproperties}
    \centering
    \begin{tabular}{|c|c|}
    \cmidrule(r){1-2}
         Embedding & Critical sketch size $m_{\delta}$  \\
        \midrule 
         SRHT & $\mathcal{O}\!\left(\log(d_e /\delta) \cdot \left(d_e +  \log(n/\delta)\right) \right)$\\
         SJLT, $s=1$& $\mathcal{O}\big(d_e^2 / \delta\big)$ \\
         %SJLT, $s=\mathcal{O}(\log(d_e/\delta))$ & $\mathcal{O}\big(d_e \log(d_e/\delta)\big)$ \\
         sub-Gaussian & $\mathcal{O}\!\left(d_e + \log(1/\delta))\right)$ \\
    \bottomrule
    \end{tabular}
\end{table}

\subsection{Approximate Newton Decrement}

Given $x \in \real^d$, the error $\delta_{x} = \frac{1}{2}\|x_t-x^*\|_H^2$ can also be expressed in its Newton decrement form~\cite{boyd2004convex}, i.e., $\delta_x = \frac{1}{2} \|\nabla f(x)\|^2_{H^{-1}}$. Due to the expensive cost $\mathcal{O}(nd^2)$ of computing $\delta_x$, we propose to approximate $\delta_x$ with
\begin{align}
    \wtilde \delta_x \defn \frac{1}{2} \nabla f(x) H_S^{-1} \nabla f(x)\,.
\end{align}
In conjunction with, for instance, the IHS that requires solving at each iteration the linear system $H_S \, v = \nabla f(x_t)$, the cost of computing $\wtilde \delta_t \equiv \wtilde \delta_{x_t}$ is negligible as it only involves an additional inner product between the available quantities $H_S^{-1} \nabla f(x_t)$ and $\nabla f(x_t)$. We have the following approximation bound between $\delta_x$ and $\wtilde \delta_x$.
\begin{lemma} 
\label{lemmaapproximatenewtondecrement}
Let $\rho \in (0,1)$. Conditional on $\mathcal{E}_\rho^m$, it holds for any $x\in \real^d$ that $|\delta_x - \wtilde \delta_x| \less \sqrt{\rho} \, \wtilde \delta_x$. Consequently, for $m > m_\delta$, we have $|\delta_x - \wtilde \delta_x| \less \sqrt{\frac{m_\delta}{m}} \,\wtilde \delta_x$ with probability $1-\delta$.
\end{lemma} 
It is also of interest to characterize the error between $\delta_x$ and $\wtilde \delta_x$ when $m \less m_\delta$.
\begin{lemma}
\label{lemmaapproximatenewtondecrementsmallsketchsize}
Let $\rho \gre 1$. Conditional on $\mathcal{E}^m_\rho$, it holds for any $x \in \real^d$ that $\delta_x \less \wtilde \delta_x \cdot (1+\rho)$. Consequently, when $m \less m_\delta$, we have $\delta_x \less \wtilde \delta_x \left(1+\frac{m_\delta}{m}\right)$ for all $x \in \real^d$ with probability $1-\delta$.
\end{lemma}

\subsection{Preconditioned First-order Methods}

\begin{definition} 
We say that the sequence of functions $\{\psi_t\}_{t \gre 0}$ is a preconditioned first-order method if given any preconditioner $P \succ 0$ and any initial point $x_0 \in \real^d$, it defines a sequence of iterates $x_{t+1} = \psi_t(x_0, \dots,x_t; P)$ such that 
\begin{align}
    x_{t+1} \in x_0 + P^{-1} \cdot \text{span}\!\left\{\nabla f(x_0), \hdots, \nabla f(x_t)\right\}\quad \mbox{for all } t\gre 0\,.
\end{align}
\end{definition} 
\begin{condition} 
\label{conditionlinearconvergence}
Given $\rho \in (0,1)$, $\phi : (0,1) \to \real_+$ and $\alpha \gre 0$, we say that a preconditioned first-order method $\{\psi_t\}_{t \gre 0}$ satisfies $(\rho,\phi(\rho), \alpha)$-linear convergence if for any $x_0 \in \real^d$, $m \gre 1$ and $S \in \real^{m \times n}$, it holds conditional on $\mathcal{E}^m_\rho$ that the sequence $x_{t+1} = \psi_t(x_0, \dots, x_t; H_S)$ satisfies
\begin{align}
    \delta_t \less \alpha \cdot \phi(\rho)^t \cdot \delta_0 \quad \mbox{for all } t\gre 0\,.
\end{align}
\end{condition} 
Under the linear convergence \cref{conditionlinearconvergence}, we obtain from \cref{lemmaapproximatenewtondecrement} a similar linear convergence guarantee for the ratio of approximate Newton decrements. We recall the notation $c(\alpha,\rho) = \frac{1+\sqrt{\rho}}{1-\sqrt{\rho}} \cdot \alpha$.
\begin{corollary}
\label{corollaryapproximateratio}
Let $\rho \in (0,1)$ and $\{\psi_t\}$ be a preconditioned first-order method with $(\rho, \phi(\rho), \alpha)$-linear convergence. Conditional on $\mathcal{E}^m_\rho$, it holds for any $x_0 \in \real^d$, $m \gre 1$ and $S \in \real^{m \times n}$ that 
\begin{align}   
    \wtilde \delta_t \less c(\alpha,\rho) \cdot \phi(\rho)^t \cdot \wtilde \delta_0\,.
\end{align}
\end{corollary}

\subsection{Proof of Results in \cref{sectionpreliminaries}}

\subsubsection{Proof of \cref{lemmaapproximatenewtondecrement}}

Let $\rho \in (0,1)$ and suppose that $\mathcal{E}^m_\rho$ holds, i.e., $\|C_S - I_d\|_2 \less \sqrt{\rho}$. Fix $x \in \real^d$. Note that $\wtilde \delta_x = \frac{1}{2} \|H_S^{-\frac{1}{2}} H^\frac{1}{2} H^{-\frac{1}{2}} \nabla f(x)\|_2^2$, and consequently, $\sigma_\text{min} (H^{\frac{1}{2}} H_S^{-1}H^{\frac{1}{2}}) \cdot \delta_x \less \wtilde \delta_x \less \sigma_\text{max}(H^{\frac{1}{2}} H_S^{-1}H^{\frac{1}{2}}) \cdot \delta_x$. Using that $H^{\frac{1}{2}} H_S^{-1}H^{\frac{1}{2}} = C_S^{-1}$, we further obtain $\frac{1}{1+\sqrt{\rho}} \cdot \delta_x \less \wtilde \delta_x \less \frac{1}{1-\sqrt{\rho}} \cdot \delta_x$, which is equivalent to the claimed result.

\subsubsection{Proof of \cref{lemmaapproximatenewtondecrementsmallsketchsize}}

Let $\rho \gre 1$ and suppose that $\mathcal{E}^m_\rho$ holds, i.e., $\|C_S - I_d\|_2 \less \rho$. Fix $x \in \real^d$. Using the same arguments as in the proof of \cref{lemmaapproximatenewtondecrement}, we obtain $\sigma_\text{min}(C_S^{-1}) \cdot \delta_x \less \wtilde \delta_x$, that is, $\delta_x \less (1+\rho)\,\wtilde \delta_x$.

\subsubsection{Proof of \cref{corollaryapproximateratio}}

Under the hypotheses of \cref{corollaryapproximateratio}, it holds that $\frac{\delta_t}{\delta_0} \less \alpha \cdot \phi(\rho)^t$. From \cref{lemmaapproximatenewtondecrement}, we have $\wtilde \delta_t \less \frac{1}{1-\sqrt{\rho}} \cdot \delta_t$ and $\wtilde \delta_0 \gre \frac{1}{1+\sqrt{\rho}} \cdot \delta_0$. Combining all these inequalities, we obtain the claimed result, i.e., $\frac{\wtilde \delta_t}{\wtilde \delta_0} \less \frac{1+\sqrt{\rho}}{1-\sqrt{\rho}} \cdot \alpha \cdot \phi(\rho)^t$.

\section{Convergence Guarantees for Preconditioned First-order Methods}
\label{sectionconvergenceanalysis}

We first develop a lower bound on the performance of any preconditioned first-order method, and we show that PCG attains this lower bound. This result is classical, and we adapt it to our sketching-based setting. It draws on connections between the theory of orthogonal polynomials and optimal iterative methods for solving linear systems of equations (see, for instance,~\cite{daniel1967conjugate, nocedal2006numerical, pedregosa2020acceleration, lacotte2020optimal} for background on this topic and other interesting applications of such connections). We recall the eigenvalue decomposition $C_S = \sum_{i=1}^d \lambda_i v_i v_i^\top$, and we introduce $\xi_i = \langle v_i, H^\frac{1}{2} (x_0 -x^*) \rangle$. We emphasize that our discussion and results in this section follow the lines of~\cite{lacotte2019faster}. In contrast to our work, they considered unregularized least-squares.
\begin{lemma}
\label{lemmalowerboundfirstordermethods}
Let $S \in \real^{m \times n}$ and $x_0\in \real^d$. It holds that any first-order method preconditioned with $H_S$ and starting at $x_0$ satisfies at each iteration
\begin{align}
\label{eqnlowerboundfirstordermethods}
    \delta_t \gre \big\{\ell^*_t(S,x_0) \defn \frac{1}{2} \min_{\substack{Q_t \in \real_t[X];\\Q_t(0)=1}} \sum_{i=1}^d Q_t(\lambda_i^{-1})^2 \xi_i^2 \big\}\,,
\end{align}
\end{lemma} 

We consider two instances of preconditioned first-order methods, namely, the IHS and PCG. We prove that both methods satisfy $(\rho, \phi(\rho), \alpha)$-linear convergence for some parameters $\phi$ and $\alpha$ and for any $\rho \in (0,1)$. 

\subsection{Iterative Hessian Sketch}

The IHS is an instance of a preconditioned first-order method with the update function $\psi_t(x_0, \dots, x_t; H_S) = x_t - \mu_t H_S^{-1} \nabla f(x_t)$. In fact, it is equivalent to the steepest descent method applied to the preconditioned quadratic objective function $f_S : z \mapsto \frac{1}{2} \|H_S^{-\frac{1}{2}} z\|_H^2 - b^\top H_S^{-\frac{1}{2}} z$. We show that the IHS satisfies $(\rho, \phi(\rho), \alpha)$-linear convergence with $\phi(\rho) = \rho$ and $\alpha=1$. The proof of this result follows from standard analysis of steepest descent methods~\cite{polyak1964some}. Similar versions of this result have already been established in~\cite{ozaslan2019regularized, lacotte2019faster}, and we recall the proof in \cref{prooftheoremrateihs} for completeness.
\begin{theorem}
\label{theoremrateihs}
Let $\rho \in (0,1)$ and consider the step sizes $\mu_t \defn 1-\rho$. Then, for any $x_0 \in \real^d$, $m \gre 1$ and $S \in \real^{m \times n}$, it holds conditional on $\mathcal{E}^m_\rho$ that
\begin{align}
\label{eqnrateihs}
    \frac{\delta_t}{\delta_0} \less \rho^t\,.
\end{align}
\end{theorem}

\subsection{Preconditioned Conjugate Gradient Method}
\label{sectionpcg}

It is well-known that PCG, as described in~\eqref{eqnpcgupdate}, is equivalent to standard CG applied to the preconditioned quadratic function $f_S : z \mapsto \frac{1}{2} \|H_S^{-\frac{1}{2}} z\|_H^2 - b^\top H_S^{-\frac{1}{2}} z$ and starting at $z_0 = H_S^{\frac{1}{2}} x_0$ (see, e.g.,~\cite{d2018optimization, axelsson1996iterative}). Precisely, CG applied to $f_S$ returns a sequence of iterates $\{z_t\}$ such that $x_t = H_S^{-\frac{1}{2}} z_t$ for all $t \gre 0$. Hence, we can leverage standard properties of CG to establish the next optimality result that we prove in details for completeness.
\begin{theorem}
\label{theoremratecg}
It holds that PCG is an instance of a preconditioned first-order method and that $\delta_t = \ell^*_t(S,x_0)$ at each iteration.
\end{theorem}
The exact error $\ell^*_t(S,x_0)$ of PCG depends on all eigenvalues of $C_S$; it seems arduous to characterize it more explicitly (e.g., in a non-variational form and in terms of the dimensions $n,d,m$) and it is expensive to compute it numerically. We will use instead the following classical upper bound (e.g., Theorem~1.2.2 in~\cite{daniel1967conjugate}) in terms of the extreme eigenvalues $\lambda_1$ and $\lambda_d$. That is, we have $\ell_t^*(S,x_0) \less 4 \cdot \left(\frac{\sqrt{\lambda_1}-\sqrt{\lambda_d}}{\sqrt{\lambda_1}+\sqrt{\lambda_d}}\right)^{2t} \cdot \delta_0$. Therefore, conditional on $\mathcal{E}^m_{\rho}$, we obtain for PCG that
\begin{align}
\label{eqnratepcg}
    \frac{\delta_t}{\delta_0} \less 4 \cdot \Big(\frac{1-\sqrt{1-\rho}}{1+\sqrt{1-\rho}}\Big)^t\,.
\end{align} 
Equivalently, conditional on $\mathcal{E}_\rho^m$, PCG satisfies $(\rho,\phi(\rho),\alpha)$ linear convergence with $\phi(\rho) = \frac{1-\sqrt{1-\rho}}{1+\sqrt{1-\rho}}$ and $\alpha=4$.
Note that the convergence rate $\phi(\rho)$ of PCG is always smaller than that of the IHS and up to four times smaller for small values of $\rho$. In fact, the upper bound~\eqref{eqnratepcg} is in general overly conservative, and finer bounds could be obtained in terms of the intermediate eigenvalues of $C_S$ (see Theorem~5.5 in~\cite{nocedal2006numerical}). This would eventually require a tight characterization of these intermediate eigenvalues and we do not pursue this direction.

\subsection{Preconditioned Chebyshev Iterations}

We consider the more general IHS update formula with heavy-ball momentum (which we refer to as the Polyak-IHS update) given by $x_{t+1} = x_t - \mu_t H_S^{-1} \nabla f(x_t) + \beta_t (x_t - x_{t-1})$. This is also known as the Chebyshev iterative method~\cite{manteuffel1977tchebychev} with preconditioner $H_S$, or the second-order Richardson method for fixed step size $\mu$ and momentum parameter $\beta$ (see, e.g., \cite{varga1961} for more background). For appropriate choices of fixed $\mu$ and $\beta$, it can be shown (see \cref{sectionchebychev} in the Supplementary Material) conditional on $\mathcal{E}_\rho^m$ that 
\begin{align}
\label{eqnasymptoticratepolyakihs}
    \limsup_{t \to \infty} \, (\delta_t / \delta_0)^{\frac{1}{t}} \less \frac{1-\sqrt{1-\rho}}{1+\sqrt{1-\rho}}\,.
\end{align}
Note that asymptotic guarantees of the form~\eqref{eqnasymptoticratepolyakihs} for heavy-ball accelerated methods are common in the literature~\cite{lessard2016analysis, loizou2020momentum, lacotte2019faster}, and this essentially comes from the approximation of the spectral radius by Gelfand's formula. 

Importantly, Polyak-IHS achieves asymptotically the same accelerated rate~\eqref{eqnratepcg} as PCG, and it is sometimes preferred to PCG in settings with high communication costs~\cite{meng2014lsrn, varga1961}. In order to develop later an adaptive version of Polyak-IHS, we would need a tight $(\rho,\phi(\rho),\alpha)$-linear convergence guarantee. We provide such a guarantee (see \cref{sectionchebychev} in the Supplementary Material) by leveraging error bounds~\cite{kozyakin2009accuracy} on the approximation of the spectral radius by Gelfand's formula. However, the value of $\alpha$ we do provide is too large to make our adaptive version of Polyak-IHS practical, and we leave as future work further investigations of an adaptive version of Polyak-IHS.

\subsection{Proofs of Results in \cref{sectionconvergenceanalysis}}

\subsubsection{Proof of \cref{lemmalowerboundfirstordermethods}}

Fix $S \in \real^{m \times n}$ and $x_0 \in \real^d$. Consider a preconditioned first-order method based on $S$ and starting at $x_0$. We show by induction that, for any $t \gre 1$, there exists $Q_t \in \real_{t}[X]$ such that $Q_t(0)=1$ and $x_t-x^* = Q_t(H_S^{-1}H) (x_0 - x^*)$. For $t=1$, we have $x_1 = x_0 + \alpha H_S^{-1} H (x_0 - x^*)$ for some $\alpha \in \real$. Setting $Q_1(X) = 1 + \alpha X$ which is a polynomial of degree $1$ yields the induction claim for $t=1$. Suppose that the induction hypothesis holds for some $t \gre 1$. We have $x_{t+1} = x_0 + H_S^{-1} \sum_{j=0}^t \alpha_j \nabla f(x_j)$ for some $\alpha_0, \dots, \alpha_t \in \real$. Note that $\nabla f(x_j) = H (x_j - x^*)$, so that $x_{t+1} = x_0 + H_S^{-1}H \sum_{j=0}^t \alpha_j (x_j - x^*)$. By induction hypothesis, for each $0 \less j \less t$, there exists $Q_j \in \real_{j}[X]$ such that $Q_j(0)=1$ and $x_j - x^* = Q_j(H_S^{-1}H) (x_0 - x^*)$. Hence, $x_{t+1} = x_0 + H_S^{-1}H\sum_{j=0}^t \alpha_j Q_j(H_S^{-1} H)(x_0 - x^*)$. Setting $Q_{t+1}(X) = 1 + X \sum_{j=0}^t \alpha_j Q_j(X)$ which is a polynomial of degree less than $t+1$ and such that $Q_{t+1}(0)=1$, we obtain $x_{t+1}-x^*=Q_{t+1}(H_S^{-1} H)(x_0 - x^*)$ and this concludes the proof of the induction hypothesis.

We now prove the lower bound~\eqref{eqnlowerboundfirstordermethods}. We fix $t \gre 1$. Let $Q_t \in \real_{t}[X]$ such that $Q_t(0)=1$ and $x_t - x^* = Q_t(H_S^{-1}H)(x_0-x^*)$. Multiplying both sides by $H^\frac{1}{2}$, we get $\Delta_t = H^\frac{1}{2} Q_t(H_S^{-1}H)(x_0-x^*)$. Note that $H^\frac{1}{2} (H_S^{-1}H)^k = (C_S^{-1})^k H^\frac{1}{2}$ for any $k \gre 0$, whence $\Delta_t = Q_t(C_S^{-1}) \Delta_0$. Consequently, $\delta_t = \frac{1}{2} (\Delta_0)^\top Q_t(C_S^{-1})^2 (\Delta_0)$. Using the eigenvalue decomposition $C_S = \sum_{i=1}^d \lambda_i v_i v_i^\top$, we find that $\delta_t = \frac{1}{2} \sum_{i=1}^d Q_t(\lambda_i^{-1})^2 \xi_i^2$. Recalling that $Q_t(0)=1$ and taking the infimum, we obtain the claimed lower bound.

\subsubsection{Proof of \cref{theoremrateihs}}
\label{prooftheoremrateihs}

Multiplying the IHS update formula~\eqref{eqnihs} by $H^\frac{1}{2}$ and subtracting $H^\frac{1}{2} x^*$, we obtain the error recursion $\Delta_{t+1} = (I_d - \mu_t C_S^{-1} ) \Delta_t$. Using that $\delta_t = \frac{1}{2} \|\Delta_t\|^2$, we obtain $\delta_{t+1} \less  \|I_d - \mu_t C_S^{-1}\|_2^2 \cdot \delta_t$, and it remains to control the operator norm of $I_d -\mu_t C_S^{-1}$. The eigenvalues of the matrix $I_d - \mu_t C_S^{-1}$ are given by $1-\frac{\mu_t}{\lambda_i}$. Then, $\|I_d - \mu_t C_S^{-1}\|_2 = \max\{|1-\frac{\mu_t}{\lambda_1}|, |1-\frac{\mu_t}{\lambda_d}|\}$. Conditional on $\mathcal{E}^m_\rho$, we have $1-\sqrt{\rho} \less \lambda_d \less \lambda_1 \less 1+\sqrt{\rho}$, whence $\max\!\left\{|1-\frac{\mu_t}{\lambda_1}|, |1-\frac{\mu_t}{\lambda_d}|\right\} \less \max\!\left\{|1-\frac{\mu_t}{1+\sqrt{\rho}}|, |1-\frac{\mu_t}{1-\sqrt{\rho}}|\right\}$. Picking $\mu_t = 1-\rho$ yields that $\max\{|1-\frac{\mu_t}{1+\sqrt{\rho}}|, |1-\frac{\mu_t}{1-\sqrt{\rho}}|\} = \sqrt{\rho}$, i.e., $\|I_d - \mu_t C_S^{-1}\|^2_2 \less \rho$, and this concludes the proof.

\subsubsection{Proof of \cref{theoremratecg}}

It is known (see, e.g.,~\cite{d2018optimization, axelsson1996iterative}) that PCG with preconditioner $H_S$ is equivalent to standard CG applied to the preconditioned quadratic function $f_S : z \mapsto \frac{1}{2} \|H_S^{-\frac{1}{2}} z\|_H^2 - b^\top H_S^{-\frac{1}{2}} z$ and starting at $z_0 = H_S^{\frac{1}{2}} x_0$. Precisely, CG applied to $f_S$ returns a sequence of iterates $\{z_t\}$ such that $x_t = H_S^{-\frac{1}{2}} z_t$ for all $t \gre 0$. We can thus leverage standard results of CG. According to Theorem~5.3 in~\cite{nocedal2006numerical}, it holds that $z_{t+1} - z_0$ lies in the span of $\{\nabla f_S(z_0), \hdots, \nabla f_S(z_{t})\}$ for all $t \gre 0$. Multiplying both terms by $H_S^{-\frac{1}{2}}$ and observing that $\nabla f_S(z_j) = H_S^{-\frac{1}{2}} \nabla f(x_j)$ for all $j \gre 0$, we obtain that $x_{t+1} \in x_0 + H_S^{-1} \mbox{span}\{\nabla f(x_0), \hdots, \nabla f(x_{t})\}$, i.e., PCG is an instance of~\eqref{eqnfirstordermethods}. According to Theorem~5.2 in~\cite{nocedal2006numerical}  (see, also, the classical references~\cite{hestenes1952methods, daniel1967conjugate}), we have
\begin{align*}
    \|H^\frac{1}{2}H_S^{-\frac{1}{2}} (z_t - z^*)\|_2^2 = \min_{\substack{Q_t \in \real_t[X];\\Q_t(0)=1}} \|Q_t(H^\frac{1}{2}H_S^{-1}H^\frac{1}{2}) H^\frac{1}{2} H_S^{-\frac{1}{2}} (z_0-z^*)\|_2^2\,. 
\end{align*}
Plugging-in the identity $x_t - x^* = H_S^{-\frac{1}{2}} (z_t - z^*)$ into the above equation, we obtain the claimed optimality result, i.e., $\delta_t = \ell_t^*(S,x_0)$.

\section{Adaptive Sketch-size First-Order Methods}
\label{sectionadaptivemethod}

Given a preconditioned first order method that satisfies, conditional on $\mathcal{E}^m_\rho$, an improvement criterion of the form $\delta_t / \delta_0 \less \alpha \cdot \phi(\rho)^t$, we propose an adaptive mechanism that tests at each iteration whether the \emph{computationally tractable} improvement test $\wtilde \delta_{t+1} / \wtilde \delta_0 \less c(\alpha,\rho) \cdot \phi(\rho)^{t+1}$ is verified. Under the hypothesis $m \gre m_\delta / \rho$, this test must hold true with probability $1-\delta$ according to \cref{corollaryapproximateratio}. Therefore, if the improvement test is not verified, we reject the hypothesis $m \gre m_\delta / \rho$: instead, we double the sketch size, sample a new embedding $S$ and restart the first-order method at $x_t$. As soon as the sketch size verifies $m \gre m_\delta / \rho$, our method is guaranteed with probability $1-\delta$ to pass the improvement test and to satisfy linear convergence. What happens under the alternative hypothesis $m < m_\delta / \rho$ if the improvement test is verified? In this case, we show that the update is still guaranteed to (approximately) make linear progress at the rate $\phi(\rho)$. This fact is crucial to our method which is then guaranteed to converge even the sketch size remains relatively small. We formally describe our prototype method in \cref{AlgorithmAdaptive} and we present its guarantees in \cref{theoremadaptive}.

\begin{algorithm}[!ht]
    \caption{Prototype Adaptive First-order Method.}
	\label{AlgorithmAdaptive}
	\begin{algorithmic}
	\STATE{\textbf{Input.} Rate $\rho \in (0,1)$, preconditioned first-order method $\{\psi_t\}_{t \gre 0}$ with $(\rho, \phi(\rho), \alpha)$ linear convergence, initial sketch size $m_\text{init} \gre 1$, initial point $x_0 \in \real^d$, and iteration number $T \gre 0$.}
	\STATE{Initialize time $t=0$, index $I=0$, and adaptive sketch size $m_0 = m_\text{init}$.}
	\STATE{Sample $S_0 \in \real^{m_0 \times n}$ and compute $\wtilde \delta_I = \frac{1}{2}\nabla f(x_0)^\top H_{S_0}^{-1} \nabla f(x_0)$.}
	\WHILE{$t < T$}
	\STATE{Compute $x^+ = \psi_{t-I}(x_I, \hdots, x_t; H_{S_t})$ and $\wtilde \delta^+ = \frac{1}{2} \nabla f(x^+)^\top H_{S_t}^{-1} \nabla f(x^+)$.}
	\IF{$\wtilde \delta^+ / \wtilde \delta_I > c(\alpha,\rho) \cdot \phi(\rho)^{t+1-I}$}
	\STATE{Set $I=t$, $m_t = 2m_t$, sample new $S_t \in \real^{m_t \times n}$ and compute $\wtilde \delta_I = \frac{1}{2} \nabla f(x_t)^\top H_{S_t}^{-1} \nabla f(x_t)$.}
	\ELSE
	\STATE{Set $x_{t+1} = x^+$, $m_{t+1} = m_t$, $S_{t+1} = S_t$, and $t = t+1$.}
	\ENDIF
	\ENDWHILE
	\end{algorithmic}
\end{algorithm}

We denote by $m_t$ the sketch size on the time interval $(t, t+1)$ (after (resp.~before) it is eventually increased at time $t$ (resp.~$t+1$)) and by $S_t \in \real^{m_t \times n}$ the corresponding sketching matrix. We denote by $K_t$ the number of times the sketch size has been increased up to time $t$.

\begin{theorem} 
\label{theoremadaptive}
Let $\rho \in (0,1/4)$ and $\{\psi_t\}_{t \gre 0}$ be a preconditioned first order method with $(\rho, \phi(\rho), \alpha)$ linear convergence. Fix $\delta \in (0,1)$ and a random embedding with critical sketch size $m_\delta$. For any $m_\text{init} \gre 1$, it holds with probability $1-(1+ K_\text{max}) \delta$ simultaneously for all $t \gre 0$ that $m_t \less \max\{m_\text{init}, 2 \, \frac{m_{\delta}}{\rho}\}$, $K_t \less K_\text{max}$ and 
\begin{align}
    \frac{\delta_t}{\delta_0} \less \alpha \cdot \phi(\rho)^{t} \cdot \Big(c(\alpha,\rho) \cdot \frac{\sigma_1^2 + \nu^2}{\nu^2}\Big)^{K_\text{max}} 2^{K_\text{max}^2}\,,
\end{align}
where $K_\text{max} = \lceil \log_2(\frac{m_{\delta}}{m_\text{init}\rho})_+ \rceil$.
\end{theorem} 
Note that the number of times the sketch size is increased is upper bounded as $K_t \less K_\text{max}$ for any $t \gre 0$, whence the algorithm terminates in less than $T+K_\text{max}$ iterations of the \emph{while} loop.

\subsection{Computational Complexity of Adaptive Methods}
\label{sectioncomplexity}

We say that an iterate $x_t$ is $(\varepsilon,\delta)$-relative accurate if $\delta_t / \delta_0 \less \varepsilon$ with probability $1-\delta$. Let $\rho \in (0,1/4)$ be a convergence rate parameter bounded away from $0$ (e.g., $\rho = 1/8$), and consider the hypotheses of \cref{theoremadaptive}. 

\subsubsection{Sketching and factorization costs} 
\label{sectionfactorization}
We denote by $\mathcal{C}^{m,n,d}_\text{sketch}$ the cost of forming $S \cdot A$ for $S \in \real^{m \times n}$. To solve \emph{exactly} a linear system of the form $H_S \cdot v = z$ at each iteration, we pre-compute and cache a matrix factorization as follows.

If $m \gre d$, we form $H_S = (SA)^\top SA + \nu^2 \Lambda$ in time $\mathcal{O}(md^2)$ and we compute its Cholesky decomposition in time $\mathcal{O}(d^3)$. Given this factorization, the cost of solving a linear system $H_S \cdot v = z$ is $\mathcal{O}(d^2)$.

If $m < d$, we form $W_S = SA \Lambda^{-1} (SA)^\top + \nu^2 I_m$ in time $\mathcal{O}(m^2 d)$, and we compute its Cholesky decomposition in time $\mathcal{O}(m^3)$. Given this factorization, the cost of solving a linear system $H_S \cdot v = z$ is $\mathcal{O}(md)$ through the Woodbury matrix identity $v = \frac{\Lambda^{-1}}{\nu^2}(I_d - (SA)^\top W_S^{-1} SA \Lambda^{-1}) z$.

Each time a new matrix $S_t$ is sampled, we compute a new sketch $S_t A$ and a new factorization of the linear system and this costs at most $\mathcal{O}(\mathcal{C}^{m_\delta,n,d}_\text{sketch} + \min\{m_\delta,d\} m_\delta d)$ because $m_t = \mathcal{O}(m_\delta)$; this occurs at most $\mathcal{O}(\log(m_\delta))$ times because $K_t = \mathcal{O}(\log(m_\delta))$. 

\subsubsection{Per-iteration cost} At each iteration, we form the gradient $\nabla f(x^+)$ in time $\mathcal{O}(nd)$, we compute the update $x^+ = \psi_{t-I}(x_I, \dots, x_t; H_{S_t})$ in time $C_\text{update}$, and we solve linear systems based on $H_{S_t}$ in time $\mathcal{O}(\min\{m,d\} d)$ which is negligible compared to $\mathcal{O}(nd)$. Hence, the per-iteration cost is $\mathcal{O}(C_\text{update} + nd)$. For the IHS and PCG, we have $C_\text{update} = \mathcal{O}(nd)$ and the per-iteration cost reduces to $\mathcal{O}(nd)$.  

\subsubsection{Total cost} According to \cref{theoremadaptive}, the total number of iterations $T$ to reach $(\varepsilon,\delta)$-accuracy scales as $T = \mathcal{O}\!\left(\log(1/\varepsilon) + \log^2(m_\delta)\right)$. Combining everything, the total cost to achieve $(\varepsilon,\delta)$-accuracy is given by
\begin{small}
\begin{align}
\label{eqntotalcomplexity}
    \mathcal{C}_{\varepsilon,\delta} = \mathcal{O}\big( (nd + \mathcal{C}_\text{update}) (\log(\frac{1}{\varepsilon}) + \log^2(m_\delta)) + \log(m_\delta) (\mathcal{C}^{m_\delta,n,d}_\text{sketch} + \min\{m_\delta,d\} m_\delta d) \big)\,.
\end{align}
\end{small}
Given $\varepsilon \in (0,1)$ and $\rho=1/8$, we summarize the space and time complexities of adaptive PCG, whose details are summarized in Algorithm \ref{algorithmadaptivepcg}, to reach an $(\varepsilon,\delta)$-accurate solution in \cref{tablecomplexities} for the SRHT, the SJLT with $s=1$ non zero entry per column and sub-Gaussian embeddings. We leave details of the proof to the reader, as it suffices to plug-in the claimed value of $m_\delta$ from \cref{tableembeddingproperties} into~\eqref{eqntotalcomplexity}.
\begin{table}[!h]
\caption{Space complexity $m_\delta$ and time complexity $\mathcal{C}_{\varepsilon,\delta}$ to reach $(\varepsilon,\delta)$-accurate solution for: adaptive PCG (\emph{Adaptive}), PCG without adaptivity and with knowledge of $d_e$ (\emph{NoAda-$d_e$}), and PCG without knowledge of $d_e$ (\emph{NoAda-$d$}). We drop the big-O notations for the sake of clarity. Note that similar complexities hold for the IHS. Here, we consider the SJLT with $s=1$ non-zero entry per column.}
  \label{tablecomplexities}
  \centering
  \begin{scriptsize}
  \begin{tabular}{|c|c|c|c|c|}
    \cmidrule(r){1-5}
     Sketch & Method & $m_\delta$ & $\mathcal{C}_{\varepsilon,\delta}$ & $\delta$\\
    \midrule
    SRHT & Adaptive & $d_e\log(d_e)$ & $nd(\log(\frac{1}{\varepsilon}) + \log^2(d_e)) + \min\{d_e\log(d_e),d\} d d_e\log^2(d_e)$ & $\frac{1}{d_e}$\\
    SRHT & NoAda-$d_e$ & $d_e \log(d_e)$ &
    $nd(\log(\frac{1}{\varepsilon}) + \log(d_e)) + \min\{d_e \log(d_e), d\} d d_e\log(d_e)$ & $\frac{1}{d_e}$\\
    SRHT & NoAda-$d$ & $d \log(d)$ &
    $nd(\log(\frac{1}{\varepsilon}) + \log(d)) + d^3 \log(d)$ & $\frac{1}{d}$\\
    \midrule
    SJLT & Adaptive & $d_e^2/\delta$ & $nd (\log(\frac{1}{\varepsilon}) + \log^2(\frac{d^2_e}{\delta})) + \min\{\frac{d_e^2}{\delta},d\}\frac{dd_e^2}{\delta}\log(d^2_e/\delta)$ & $\delta$\\
    SJLT & NoAda-$d_e$ & $d_e^2/\delta$ & $nd\log(\frac{1}{\varepsilon}) + \min\{\frac{d_e^2}{\delta}, d\} \frac{d d^2_e}{\delta}$ & $\delta$\\
    SJLT & NoAda-$d$ & $d^2/\delta$ & $nd\log(\frac{1}{\varepsilon}) + \frac{d^4}{\delta}$ & $\delta$\\
    \midrule
    sub-G. & Adaptive & $d_e$ & $nd (\log(\frac{1}{\varepsilon}) + d_e \log(d_e))$ & $e^{-d_e}$\\
    sub-G. & NoAda-$d_e$ & $d_e$ & $nd (\log(\frac{1}{\varepsilon}) + d_e)$ & $e^{-d_e}$ \\
    sub-G. & NoAda-$d$ & $d$ & $nd (\log(\frac{1}{\varepsilon}) + d)$ & $e^{-d}$ \\
    \bottomrule
    \end{tabular}
    \end{scriptsize}
\end{table}

\begin{algorithm}[!ht]
    \caption{Adaptive Preconditioned Conjugate Gradient Method.}
	\label{algorithmadaptivepcg}
	\begin{algorithmic}
	\STATE{\textbf{Input.} Rate $\rho \in (0,1)$, initial sketch size $m_\text{init} \gre 1$, initial point $x_0 \in \real^d$, and iteration number $T \gre 0$. Set $\phi(\rho) = \frac{1-\sqrt{1-\rho}}{1+\sqrt{1-\rho}}$ and $c(\rho) = 4\, \frac{1+\sqrt{\rho}}{1-\sqrt{\rho}}$. Initialize time $t=0$, index $I=0$, and adaptive sketch size $m_0 = m_\text{init}$.}
	\STATE{Sample $S_0 \in \real^{m_0 \times n}$, form $S_0 A$, and compute and cache a matrix factorization of $H_{S_0}$ as described in \cref{sectionfactorization}. Compute $r_0 = b - Hx_0$, solve $H_{S_0} \wtilde r_0 = r_0$, set $p_0 = \wtilde r_0$, compute $\wtilde \delta_0 = r_0^\top \wtilde r_0$ and set $\wtilde \delta_I = \wtilde \delta_0$.}
	\WHILE{$t < T$}
	\STATE{Compute $\alpha_t = \wtilde \delta_t / p_t^\top H p_t$, $x^+ = x_t + \alpha_t p_t$, $r^+ = r_t - \alpha_t H p_t$, and solve $H_{S_t} \wtilde{r}^+ = r^+$. Compute $\wtilde \delta^+ = (r^+)^\top \wtilde{r}^+$ and $p^+ = \wtilde{r}^+ + (\wtilde \delta^+/\wtilde \delta_t) p_t$.}
	\IF{$\wtilde \delta^+ / \wtilde \delta_I > c(\rho) \cdot \phi(\rho)^{t+1-I}$}
	\STATE{Set $I=t$ and $m_t = 2m_t$. Sample $S_t \in \real^{m_t \times n}$, form $S_t A$, and compute and cache a matrix factorization of $H_{S_t}$ as described in \cref{sectionfactorization}. Compute $r_t = b - H x_t$, solve $H_{S_t} \wtilde r_t = r_t$, set $p_t = \wtilde r_t$, compute $\wtilde \delta_t = r_t^\top \wtilde r_t$ and set $\wtilde \delta_I=\wtilde \delta_t$.}
	\ELSE
	\STATE{Set $x_{t+1} = x^+$, $r_{t+1} = r^+$, $\wtilde r_{t+1} = \wtilde r^+$, $p_{t+1} = p^+$, $\wtilde \delta_{t+1} = \wtilde \delta^+$, $m_{t+1} = m_t$, $S_{t+1} = S_t$, and $t = t+1$.}
	\ENDIF
	\ENDWHILE
	\end{algorithmic}
\end{algorithm}

\begin{remark}
\label{remarkterminationcriterion}
We discuss here a simple termination criterion in terms of the approximate Newton decrement that guarantees $(\varepsilon, \delta)$-accuracy of an iterate $x_t$. Let $m \gre 1$ be any sketch size and consider the termination criterion 
\begin{align}
\label{eqnterminationcriterion}
    \wtilde \delta_t \less \varepsilon / (m_\delta+1)\,.
\end{align}
As an immediate consequence of \cref{lemmaapproximatenewtondecrementsmallsketchsize}, it holds under~\eqref{eqnterminationcriterion} that $\delta_t \less \varepsilon$ with probability at least $1-\delta$. The termination criterion~\eqref{eqnterminationcriterion} is conservative in the sense that it assumes that the sketch size remains small compared to $m_\delta$. Thus, it incurs an additional number of iterations $\log(m_\delta)$. More importantly, the critical sketch size $m_\delta$ depends itself on $d_e$. Therefore, in practice, one may choose to replace $d_e$ by $d$ in the expression of $m_\delta$ and the additional number of iterations gets even larger. Alternatively, one can simply use as a termination criterion the objective value $f(x_t)$, but this leaves a theoretical gap with the measure of error $\delta_t = \frac{1}{2}\|x_t-x^*\|_H^2$ we use in our theoretical guarantees because $\delta_t \neq f(x_t)$ except for consistent linear systems.
\end{remark}

\subsection{Proofs of Results in \cref{sectionadaptivemethod}}

\subsubsection{Proof of \cref{theoremadaptive}}

We use the shorthand $m_{\rho,\delta} = m_{\delta}/\rho$. By definition of $m_\delta$, we have the $(\rho, \delta)$-approximation guarantee $\|C_S - I_d\|_2 \less \sqrt{\rho}$ with probability $1-\delta$ for $m \gre m_{\rho,\delta}$. We assume that $m_\text{init} < m_{\rho,\delta}$; otherwise, the claim immediately follows from the $(\rho, \phi(\rho), \alpha)$-linear convergence assumption and from \cref{corollaryapproximateratio}. We denote by $T_0$ the first time index when the sketch size becomes greater or equal to $m_{\rho,\delta}$ (set $T_0 = +\infty$ if it does not exist). We denote by $0 \less i_1 < i_2 < \hdots < i_K = T_0$ the time indices when the sketch size gets increased, and we set $i_0=0$. Note that $\{m_t\}$ is a non-decreasing sequence, that $\frac{m_{i_j}}{m_\text{init}} \gre 2^j$ for $j=0,\dots,K$, that $m_{i_{K-1}} < m_{\rho,\delta}$ and $m_{\delta,\rho} \less m_{i_K} < 2 m_{\delta,\rho}$, whence $K \less \lceil \log_2(\frac{m_{\rho,\delta}}{m_\text{init}}) \rceil$.

We claim that, conditional on $T_0 < +\infty$, it holds with probability $1-\delta$ that $m_t = m_{T_0}$ for all $t \gre T_0$ (whence $m_t < 2 m_{\rho,\delta}$) and that $\delta_t / \delta_{T_0} \less \alpha \cdot \phi(\rho)^{t-T_0}$. Indeed, suppose that $\|C_{S_{T_0}} - I_d\|_2 \less \sqrt{\rho}$. Note that this event holds with probability $1-\delta$ since $m_{T_0} \gre m_{\rho,\delta}$. By the $(\rho, \phi(\rho), \alpha)$-linear convergence assumption, we have $\frac{\delta_{T_0+1}}{\delta_{T_0}} \less \alpha \cdot \phi(\rho)$. From \cref{corollaryapproximateratio}, we further obtain that $\frac{\wtilde \delta_{T_0+1}}{\wtilde \delta_{T_0}} \less c(\alpha,\rho) \phi(\rho)$. Hence, the iterate is accepted and $m_{T_0+1} = m_{T_0}$. Repeating this argument, we obtain the claim by induction.

Fix $j \in \{0,\dots,K-1\}$. An elementary calculation yields that $\frac{\delta_{i_{j+1}}}{\delta_{i_j}} \less \frac{\wtilde \delta_{i_{j+1}}}{\wtilde \delta_{i_j}} \cdot \kappa(C_{S_{i_j}})$. Between the time indices $i_j$ and $i_{j+1}$, there is no rejected update whence $\frac{\wtilde \delta_{i_{j+1}}}{\wtilde \delta_{i_j}} \less c(\alpha,\rho) \phi(\rho)^{i_{j+1}-i_j}$. According to \cref{lemmaapproximatenewtondecrementsmallsketchsize}, it holds with probability $1-\delta$ that $\kappa(C_{S_{i_j}}) \less \frac{1+\frac{m_{\rho,\delta}}{m_{i_j}}}{\frac{\nu^2}{\sigma_1^2 + \nu^2}} \less \frac{\sigma_1^2 + \nu^2}{\nu^2} \cdot 2^{\log_2(\frac{m_{\rho,\delta}}{m_\text{init}}) - j+1}$. Combining these observations and using a union bound, we obtain
\begin{align*}
    \frac{\delta_{t}}{\delta_0} \less \phi(\rho)^t \Big(c(\alpha,\rho) \cdot \frac{\sigma_1^2 + \nu^2}{\nu^2}\Big)^{\lceil \log_2(\frac{m_{\rho,\delta}}{m_\text{init}}) \rceil} 2^{\lceil \log_2^2(\frac{m_{\rho,\delta}}{m_\text{init}})\rceil}\,,
\end{align*}
with probability $1-\lceil \log_2(\frac{m_{\rho,\delta}}{m_\text{init}}) \rceil \delta$ simultaneously for all $t \less T_0$. Recall that $K_\text{max} = \lceil \log_2(\frac{m_{\rho,\delta}}{m_\text{init}})\rceil$. Combining everything, we obtain with probability $1-(1+K_\text{max}) \delta$ simultaneously for all $t \gre 0$ that $m_t < 2 \cdot m_{\rho,\delta}$, $K_t \less K_\text{max}$ and 
\begin{align*}
    \frac{\delta_t}{\delta_0} \less \alpha \cdot \phi(\rho)^{t} \cdot\Big(c(\alpha,\rho) \cdot \frac{\sigma_1^2 + \nu^2}{\nu^2}\Big)^{K_\text{max}} 2^{K_\text{max}^2}\,,
\end{align*}
which is the claimed result.

%\subsubsection{Proof of Theorem~\cref{theoremcomplexityadaptive}}We plug the values of $m_\delta$ as given in Table~\cref{tableembeddingproperties} and $\mathcal{C}_\text{sketch}^{m_\delta,n,d}$ into~\eqref{eqntotalcomplexity}. For the SRHT, we pick $\delta = \mathcal{O}(1/(d_e \log d_e))$, so that $m_\delta = \mathcal{O}(d_e \log(d_e))$ and $\mathcal{C}^{m_\delta,n,d}_\text{sketch} = \mathcal{O}(nd \log d_e)$. This yields success probability $1-\mathcal{O}(1/d_e)$. For the SJLT with $s=1$ non-zero entry per column, we have $m_\delta = \mathcal{O}(d^2_e/\delta)$ and $\mathcal{C}^{m_\delta,n,d}_\text{sketch} = \mathcal{O}(nd)$. This yields success probability $1-\delta$. For sub-gaussian embeddings, we pick $\delta = \mathcal{O}(e^{-d_e})$ and we have $m_\delta = \mathcal{O}(d_e)$ and $\mathcal{C}^{m_\delta,n,d}_\text{sketch} = \mathcal{O}(ndd_e)$. This yields success probability $1-\mathcal{O}(e^{-d_e})$. %For the SJLT with $s=\log(d_e)$ non-zero entries per column, we pick $\delta = \mathcal{O}(1/(d_e \log(d_e)))$, and we have $m_\delta = \mathcal{O}(d_e \log(d_e))$ and $\mathcal{C}^{m_\delta,n,d}_\text{sketch} = \mathcal{O}(\text{nnz}(A) \log(d_e))$. This yields success probability $1-\mathcal{O}(1/d_e)$. Note that the extra number of iterations $\mathcal{O}(nd \log^2 d_e)$ induced by adaptivity dominates the contribution of sketching $\mathcal{O}(\text{nnz}(A) \log^2(d_e))$, so that the final complexity does not depend on $\text{nnz}(A)$. 

\section{Subspace Embedding Properties}
\label{sectionconcentration}

The upper bounds on $m_\delta$ we provide in \cref{tableembeddingproperties} do not specify the numerical constants involved. We aim here to provide sharp characterization of these numerical constants for the SRHT as well as for Gaussian embeddings for which one should expect to have finer bounds compared to sub-Gaussian ones. We defer the proofs of \cref{theoremconcentrationshrt} and \cref{theoremgaussianconcentration} to \cref{sectionconcentrationproofs}.
\begin{theorem} 
\label{theoremconcentrationshrt} 
Let $S \in \real^{m \times n}$ be an SRHT. It holds that $\mathbb{P}(\mathcal{E}^m_\rho) \gre 1-\delta$ provided that $m \gre m_\delta/\rho$ where
\begin{align}
    m_\delta = 16 \cdot \log(16d_e /\delta) \cdot \big(\sqrt{d_e} + \sqrt{8 \log(2n/\delta)}\big)^2\,.
\end{align}
\end{theorem} 
Given a set $\mathcal{C} \subset \real^d$, we define its Gaussian width as $\omega(\mathcal{C}) = \mathbb{E}\!\left\{\sup_{u \in \mathcal{C}}|h^\top u|\right\}$ where the expectation is taken over the random Gaussian vector $h \sim \mathcal{N}(0,I_d)$.
\begin{theorem} 
\label{theoremgaussianconcentration}
Let $S \in \real^{m \times n}$ be a matrix with i.i.d.~Gaussian entries $\mathcal{N}(0,1/m)$. Let $\mathcal{C} \subset \real^d$ be a closed set with unit radius, i.e., $\sup_{u \in \mathcal{C}}\|u\|_2 = 1$. Given $\delta > 0$ and $\rho > 0$, it holds with probability at least $1-\delta$ that
\begin{align}
    & \sup_{u \in \mathcal{C}}\, \langle u, (U^\top S^\top S U - I_d) u \rangle \less 2 \sqrt{\rho} + \rho\,,\\ 
    & \inf_{u \in \mathcal{C}}\, \langle u, (U^\top S^\top SU - I_d) u\rangle \gre -\max\big\{2 \sqrt{\rho} - \rho,\, \rho\big\}\,,
\end{align}
provided that $m \gre m_\delta / \rho$, where $m_\delta = (\omega(\mathcal{C}) + \sqrt{8\log(16/\delta)})^2$.
\end{theorem} 
Setting $\mathcal{C} = \big\{\frac{D}{\|D\|_2} \cdot x \mid \|x\|_2 \less 1\big\}$, we have $\omega(\mathcal{C})^2 \less d_e$. For $\rho \in (0,1)$, we obtain 
\begin{align}
\label{eqngaussianembeddingproperties}
     \|D\|_2^2 \cdot (-2 \sqrt{\rho} + \rho) \less \lambda_\text{min}(C_S - I_d) \less \lambda_\text{max}(C_S - I_d) \less \|D\|_2^2 \cdot (2 \sqrt{\rho} + \rho)\,, 
\end{align}
with probability $1-\delta$ provided that $m \gre (\sqrt{d_e} + \sqrt{8\log(16/\delta)})^2 / \rho$. Hence, the bounds~\eqref{eqngaussianembeddingproperties} provide a sharp characterization of the numerical constants, in contrast to those presented in \cref{tableembeddingproperties} for sub-gaussian embeddings.

Besides controlling the extreme eigenvalues of $C_S - I_d$, \cref{theoremgaussianconcentration} has also interesting implications for covariance matrix estimation: we let $X_1, \dots, X_m \in \real^d$ be independent Gaussian vectors with zero-mean and covariance matrix $\Sigma$, and we denote $\wtilde \Sigma = m^{-1} \sum_{i=1}^m X_i X_i^\top$ the empirical covariance matrix. Note that $\wtilde \Sigma - \Sigma \overset{\mathrm{d}}{=} \Sigma^\frac{1}{2} (U^\top S^\top S U - I_d) \Sigma^\frac{1}{2}$. Setting $\mathcal{C} = \big\{\frac{\Sigma^\frac{1}{2}}{\|\Sigma^\frac{1}{2}\|_2} \cdot x \mid \|x\|_2 \less 1 \big\}$, the following result is an immediate consequence of \cref{theoremgaussianconcentration}.
\begin{theorem}[Covariance matrix estimation]
\label{theoremcovariancematrixestimation}
Let $\delta \in (0,1)$ and $\rho > 0$. It holds with probability at least $1-\delta$ that
\begin{align}
     &\sup_{\|x\|_2 \less 1} x^\top (\wtilde \Sigma - \Sigma) x \less \|\Sigma\|_2 \cdot (2\sqrt{\rho} + \rho)\\
     &\inf_{\|x\|_2 \less 1} x^\top (\wtilde \Sigma - \Sigma) x \gre - \|\Sigma\|_2 \cdot \max\{2\sqrt{\rho} -\rho, \rho\}\,,
\end{align}
provided that $m \gre \frac{1}{\rho} \cdot ( \sqrt{d_\Sigma} + \sqrt{8 \log(16/\delta)})^2$, where $d_\Sigma = \frac{\|\Sigma^\frac{1}{2}\|_F^2}{\|\Sigma^\frac{1}{2}\|^2_2}$.
\end{theorem}

\section{Numerical Experiments}
\label{sectionnumericalexperiments}

We carry out a numerical evaluation of adaptive PCG (\cref{algorithmadaptivepcg}) and adaptive IHS. As for the baselines, we use a direct method with Cholesky decomposition for exact solving of the linear system, standard CG (without preconditioning), and PCG with default sketch size $m=2d$. Without prior knowledge of the effective dimension, the latter is a standard sketching-based iterative solver for large-scale and ill-conditioned problems~\cite{rokhlin2008fast, meng2014lsrn}. We implement\footnote{Our code is publicly available at \url{https://github.com/pilancilab/adaptive_effdim_solver}.} the different solvers in Python using its standard numerical linear algebra modules. For fairness of comparison, we implement each of the iterative methods considered here. We run our simulations on a computational node with $3$ Tb of memory and $64$ CPUs. The datasets that we consider fit into memory, whence we consider here the SJLT and the SRHT for their small sketching cost in this standard computational setup in comparison to sub-Gaussian embeddings. Note that in a distributed setting that we do not address here, sub-Gaussian embeddings may be preferable~\cite{meng2014lsrn}.

\begin{figure}[!ht]
	\centering
	\includegraphics[width=0.9\textwidth]{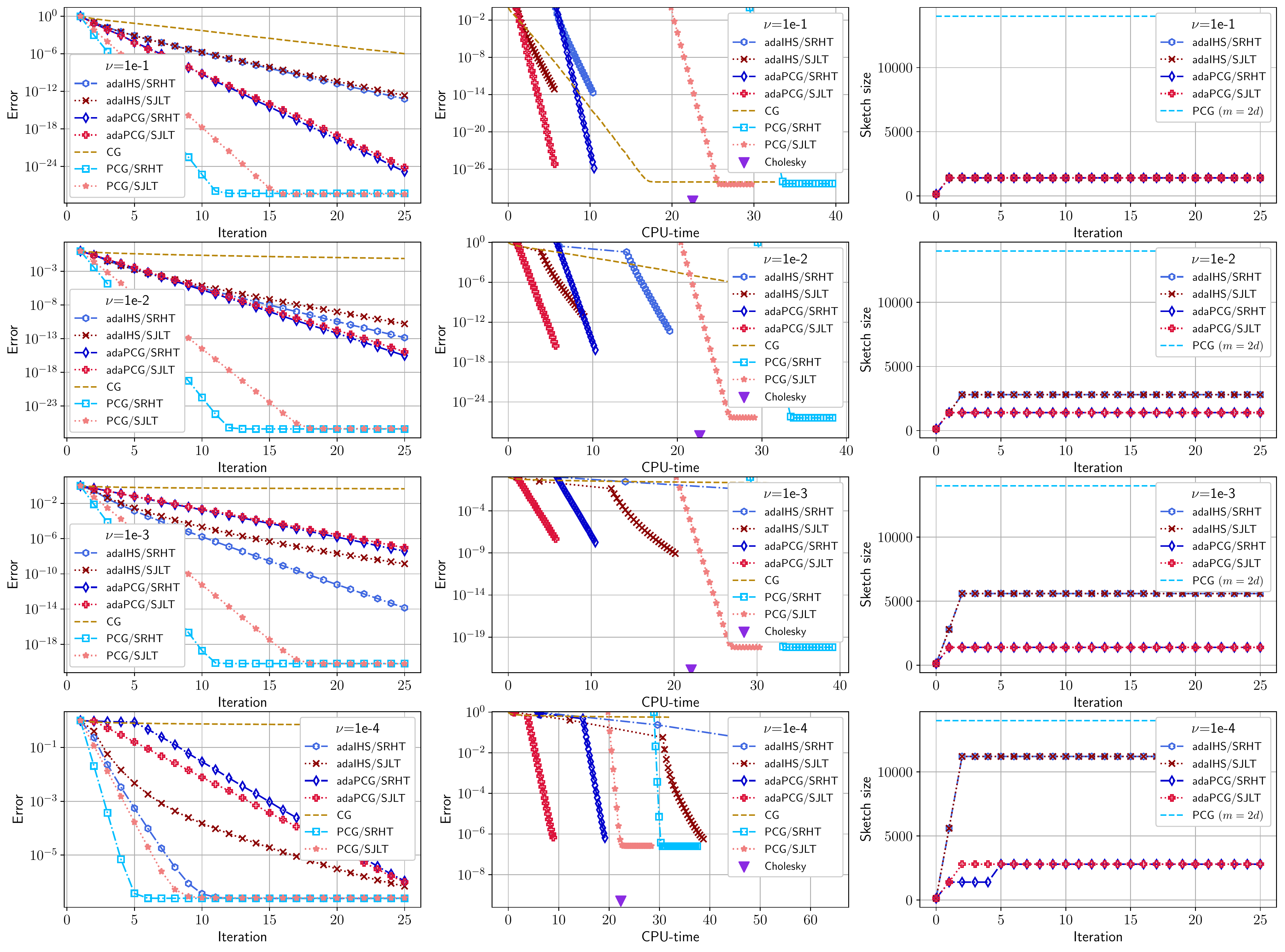}
	\caption{Synthetic dataset with $n=16384$, $d=7000$ and $\nu \in \{10^{-1}, 10^{-2}, 10^{-3}, 10^{-4}\}$, which yields respectively the effective dimensions $d_e \approx 200$, $400$, $800$, $1600$. Left column: relative error $\delta_t /\delta_0$ versus iteration number; center column: relative error $\delta_t/\delta_0$ versus CPU-time in seconds; right column: adaptive sketch size versus iteration number.}
	\label{figuresynthetic1}
\end{figure}

\begin{figure}[!ht]
	\centering
	\includegraphics[width=0.9\textwidth]{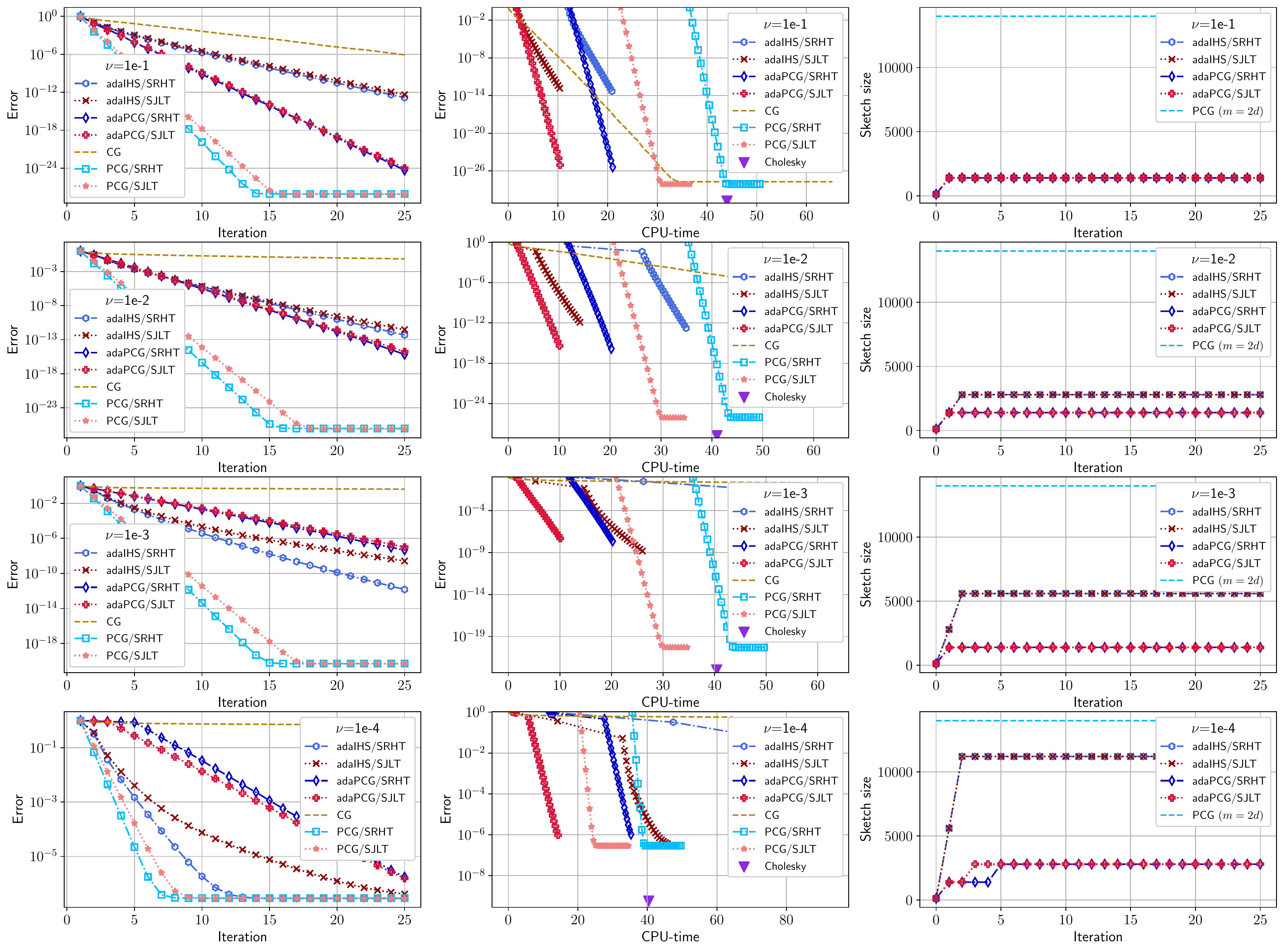}
	\caption{Synthetic dataset with $n=131072$, $d=7000$ and $\nu \in \{10^{-1}, 10^{-2}, 10^{-3}, 10^{-4}\}$, which yields respectively the effective dimensions $d_e \approx 200$, $400$, $800$, $1600$.}
	\label{figuresynthetic2}
\end{figure}

\begin{figure}[!ht]
	\centering
	\includegraphics[width=0.9\textwidth]{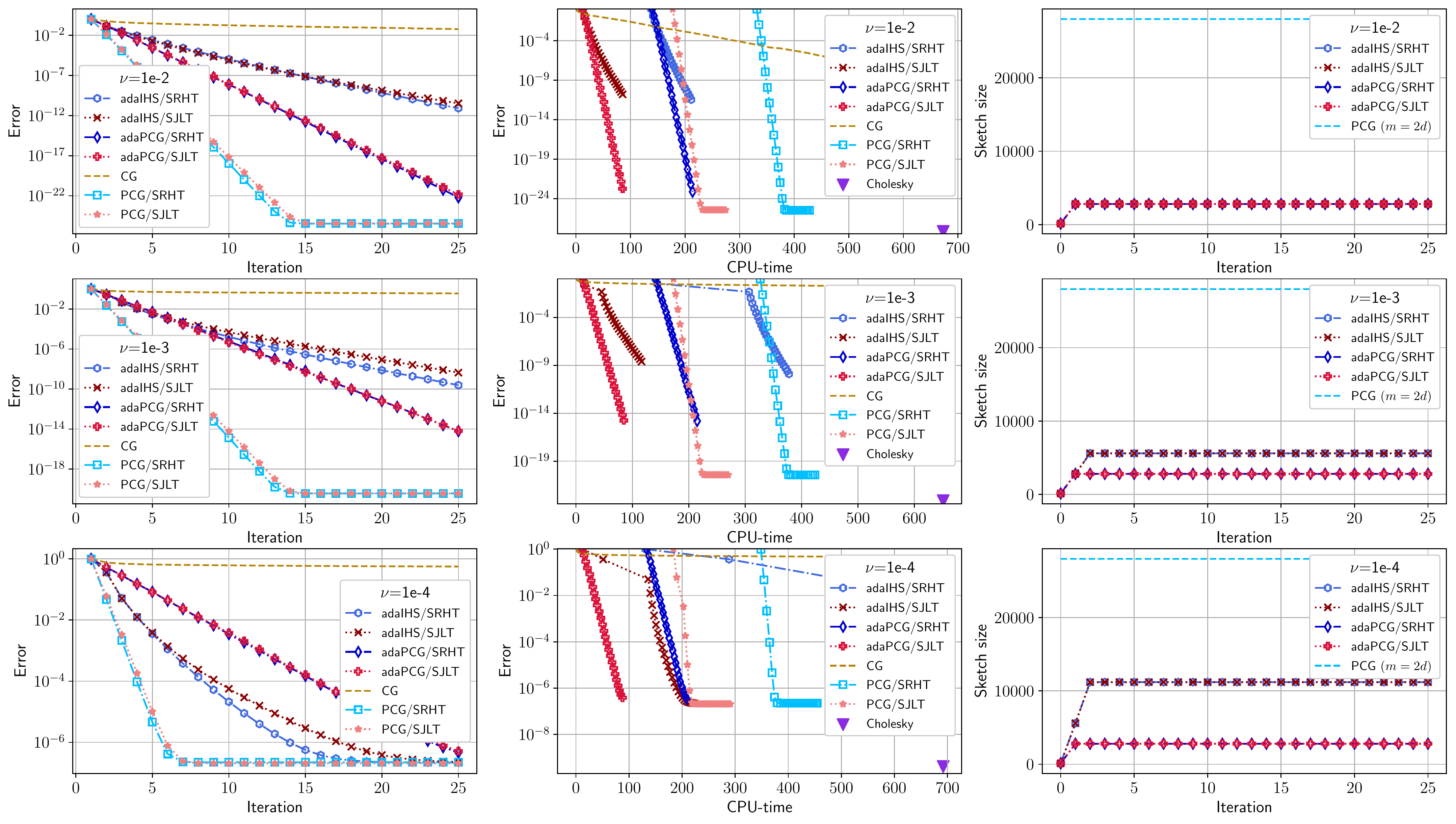}
	\caption{Synthetic dataset with $n=524288$, $d=14000$ and $\nu \in \{10^{-2}, 10^{-3}, 10^{-4}\}$, which yields respectively the effective dimensions $d_e \approx 400$, $800$, $1600$.}
	\label{figuresynthetic3}
\end{figure}

\begin{figure}[!ht]
	\centering
	\includegraphics[width=0.9\textwidth]{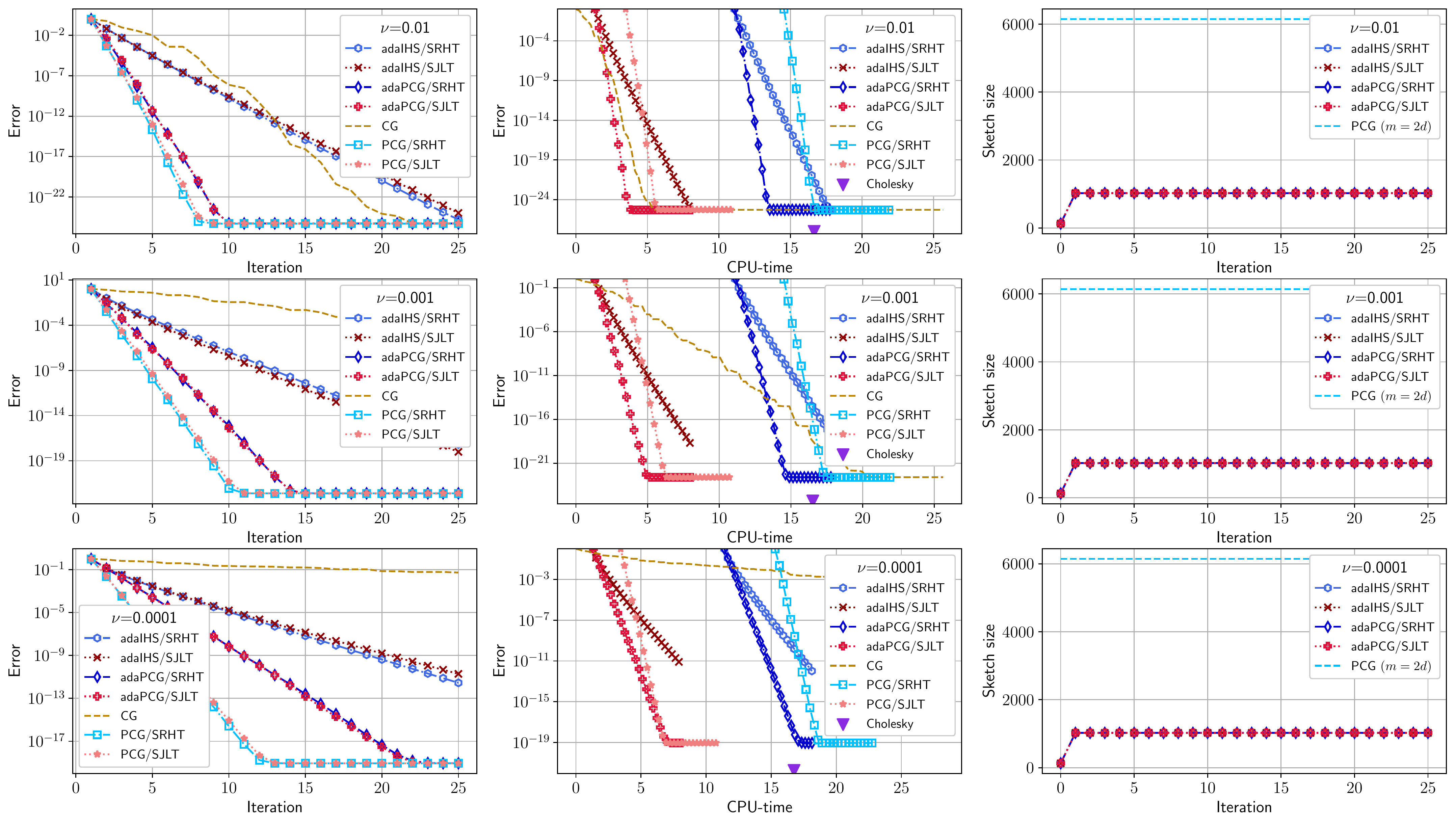}
	\caption{CIFAR-$100$ dataset: $n=60000$, $d=3073$, $c=100$ classes.}
	\label{figurecifar}
\end{figure}

\begin{figure}[!ht]
	\centering
	\includegraphics[width=0.9\textwidth]{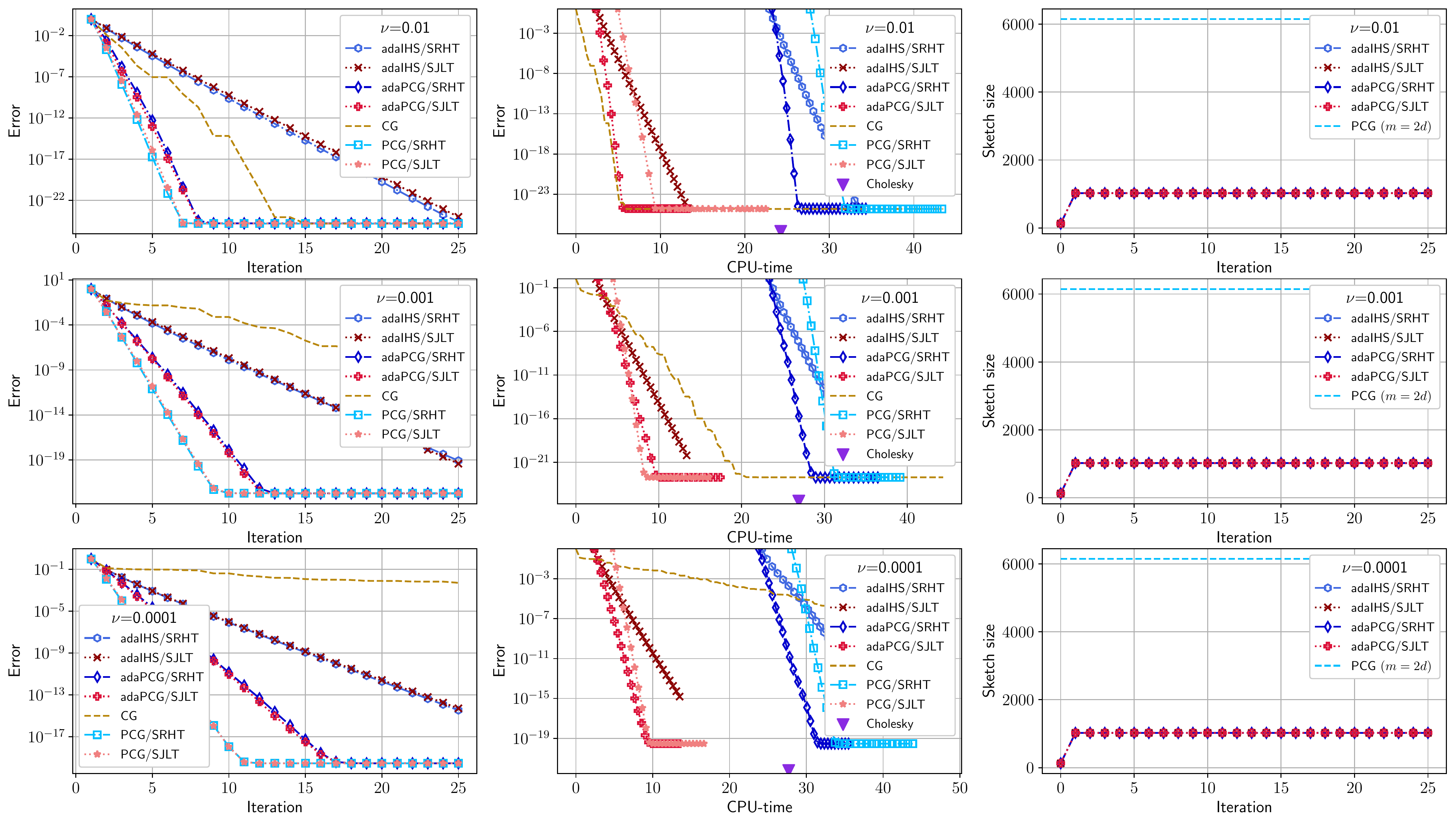}
	\caption{SVHN dataset: $n=99289$, $d=3073$, $c=10$ classes.}
	\label{figuresvhn}
\end{figure}

\begin{figure}[!ht]
	\centering
	\includegraphics[width=0.9\textwidth]{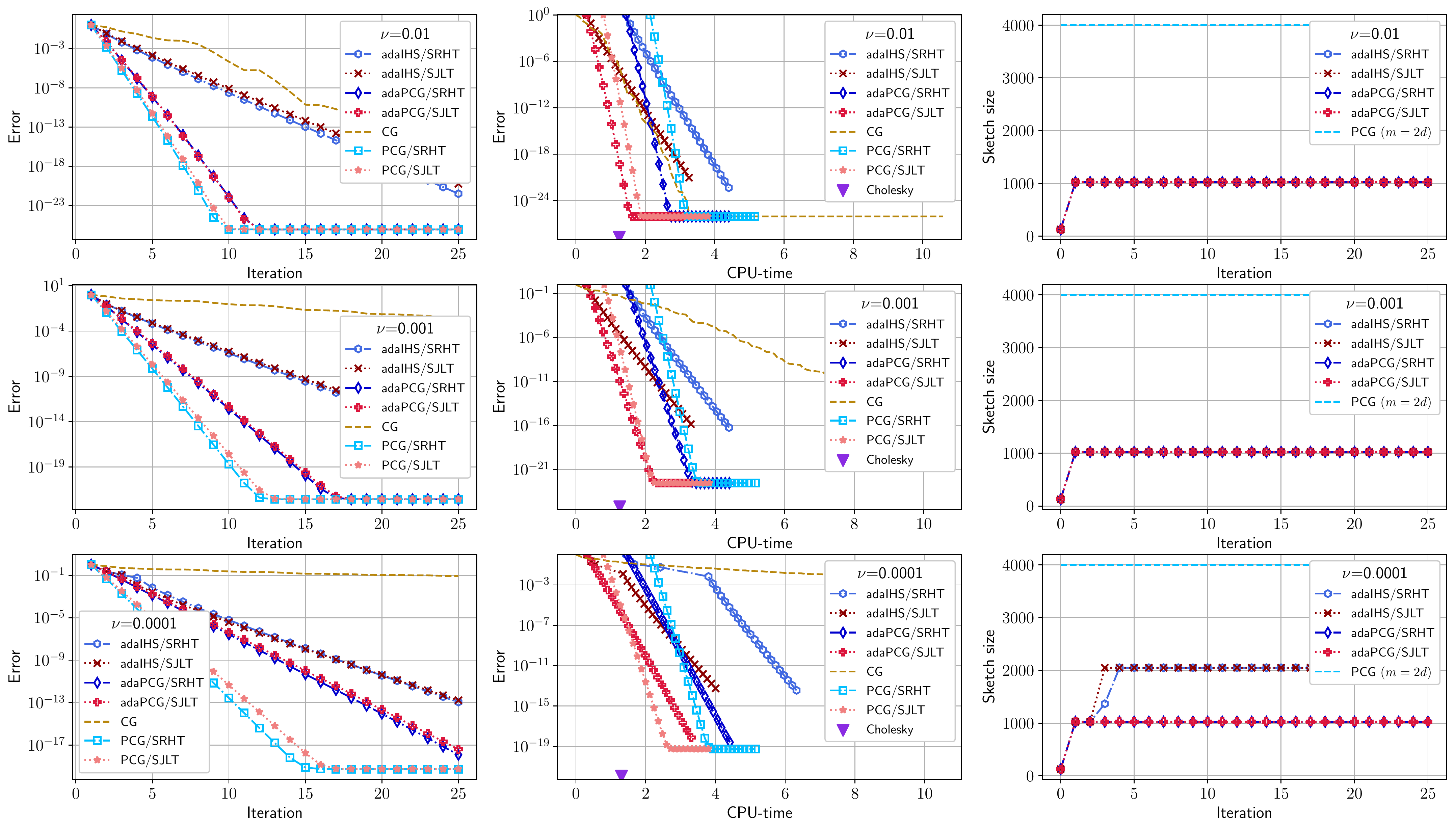}
	\caption{Dilbert dataset: $n=10000$, $d=2001$, $c=5$ classes.}
	\label{figuredilbert}
\end{figure}

\begin{figure}[!ht]
	\centering
	\includegraphics[width=0.9\textwidth]{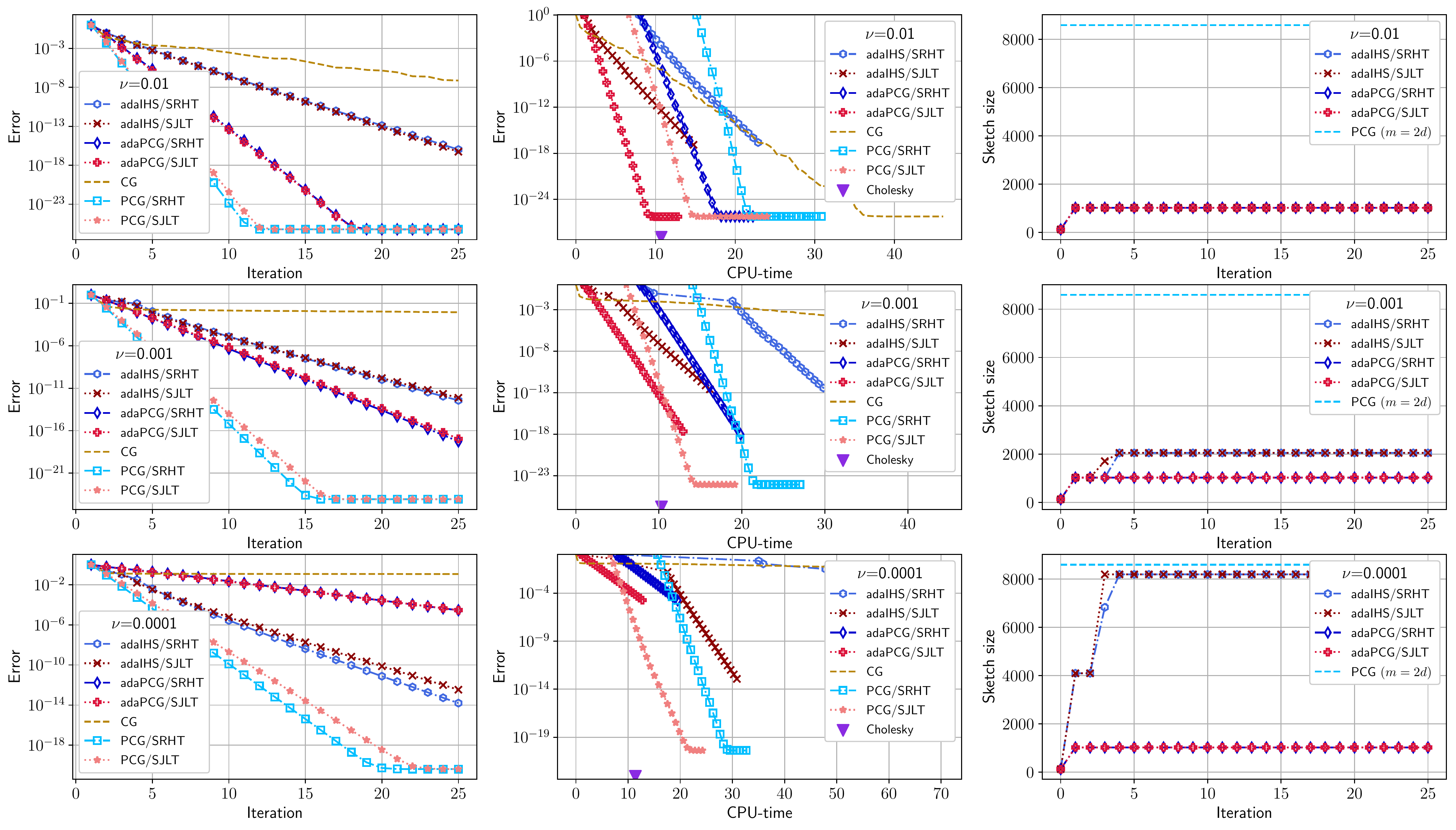}
	\caption{Guillermo dataset: $n=20000$, $d=4297$, $c=2$ classes.}
	\label{figureguillermo}
\end{figure}

\begin{figure}[!ht]
	\centering
	\includegraphics[width=0.9\textwidth]{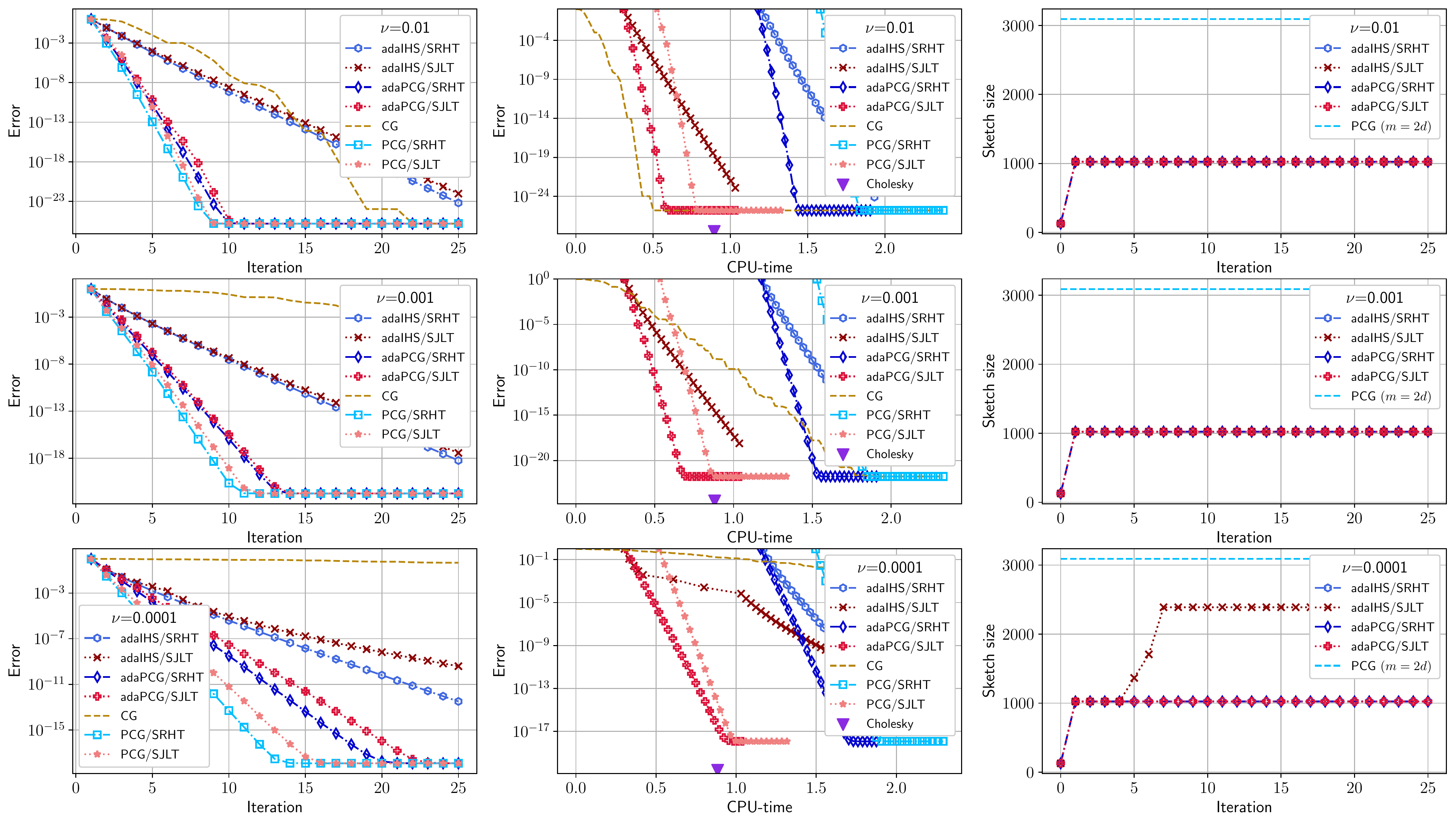}
	\caption{OVA-Lung dataset: $n=1545$, $d=10936$ and $c=2$ classes.}
	\label{figureovalung}
\end{figure}

\begin{figure}[!ht]
	\centering
	\includegraphics[width=0.9\textwidth]{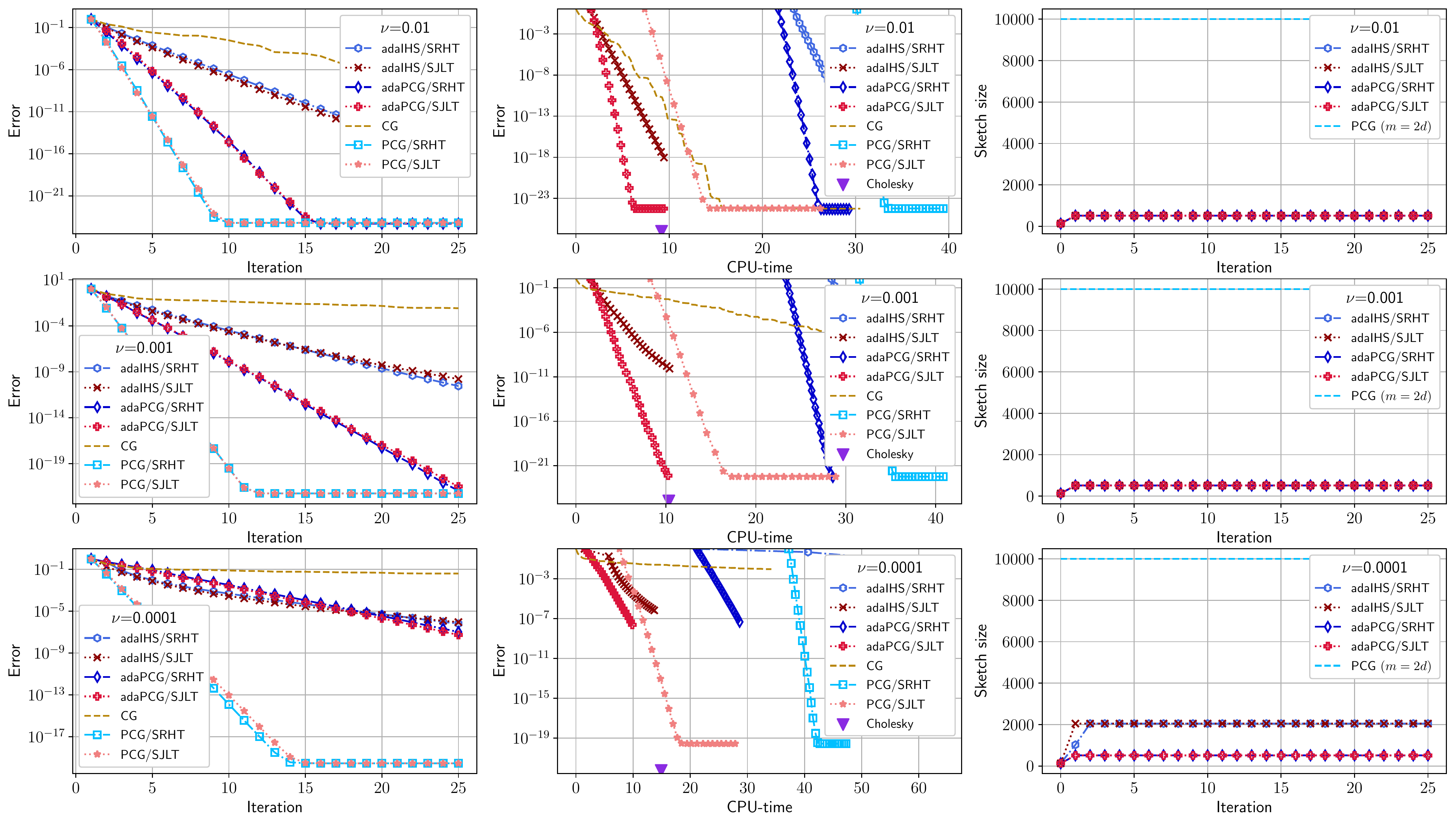}
	\caption{WESAD dataset: $n=250000$, $d=10000$ and $c=2$ classes.}
	\label{figurewesad}
\end{figure}

\subsubsection*{Synthetic datasets} We consider several synthetic datasets where the matrix $A \in \real^{n \times d}$ has singular values with exponential decay, i.e., $\sigma_j = 0.995^j$ for $j \in \{1,\dots,d\}$, and for varying dimensions $(n,d)$. For each pair $(n,d)$, we consider several regularization parameters $\nu$, i.e., several effective dimensions and condition numbers. We report results on \cref{figuresynthetic1}, \cref{figuresynthetic2} and \cref{figuresynthetic3}. We observe across all datasets that, as the regularization parameter decreases, that is, as the condition number increases, CG converges (as expected) at a slower rate. The performance of PCG is not sensitive to the condition number, and it is faster here with the SJLT in comparison to the SRHT. On the other hand, our adaptive methods offer stronger performance for most instances of these problems in terms of time and memory complexities. For the largest regularization parameter values, the effective dimension is small, and the adaptive methods starting at $m=1$ converge fast with small final sketch size. As the regularization parameter increases, the final sketch size increases. Importantly, the final sketch size always remains relatively small compared to $m=2d$, and it is considerably smaller for small effective dimensions. Hence, adaptive PCG offers significant time improvements over its effective dimension oblivious counterpart. The fastest performance is also obtained here with the SJLT. Note that the performance of adaptive IHS is weaker than that of adaptive PCG, and this is expected due to the optimality of PCG among the class of preconditioned first-order methods. In fact, adaptive IHS improves on CG and PCG with $m=2d$ over fewer instances, namely, those with large enough condition numbers and small enough effective dimensions.

In summary, adaptive PCG offers significant time and memory improvements for solving ill-conditioned linear systems in comparison to standard iterative solvers without knowledge of the effective dimension, and importantly, over a direct factorization method.

\subsubsection*{Real datasets} We aim to confirm these numerical observations over several real and open-source datasets. Except for the WESAD dataset \cite{schmidt2018introducing}, all datasets are downloaded from \url{https://www.openml.org/} with no data pre-processing. For the WESAD dataset, we use the data obtained from the E4 Empatica device and we filter the data over windows of one second\footnote{See the related Github repository \url{https://github.com/WJMatthew/WESAD/blob/master/data_wrangling.py}.}. Then we apply a random features map that approximates the Gaussian kernel $\exp(-\gamma x^2)$ with bandwidth $\gamma = 0.01$ and $d=10000$ components. Results are reported on \cref{figurecifar}, \cref{figuresvhn}, \cref{figuredilbert}, \cref{figureguillermo}, \cref{figureovalung} and \cref{figurewesad}. Note that there are datasets with more than $c=2$ classes (e.g., CIFAR-100 with $c=100$ classes): we transform the vector of labels into a hot-encoding matrix. Our implementation accounts for matrix variables, and it is straightforward to extend our theoretical results to the matrix case that, for the sake of conciseness, we do not detail here. 

Similar qualitative observations as for the synthetic datasets hold, and adaptive PCG with the SJLT often yields the best performance in terms of CPU-time for most datasets and regularization parameters.

\section{Proof of Results in \cref{sectionconcentration}}
\label{sectionconcentrationproofs}

\subsection{Proof of \cref{theoremconcentrationshrt}}
\label{sectionconcentrationsrht}

Let $S \in \real^{m \times n}$ be a SRHT matrix, that is, $S = R H \text{diag}(\varepsilon)$ where $R$ is a row-subsampling matrix of size $m \times n$, $H$ is the normalized Walsh-Hadamard transform of size $n \times n$ and $\varepsilon$ is a vector of $n$ independent Rademacher variables. We establish the following matrix deviation inequality.
Our proof follows steps closely similar to the proof of J.~Tropp~\cite{tropp2011improved} who shows concentration of the matrix $U^\top S^\top S U$ around the identity matrix $I_d$. We start with proving in \cref{lemmarownorms} that the matrix $H \text{diag}(\varepsilon) D$ preserves row norms, i.e., $\max_{j \in [n]} \|e_j^\top H \text{diag}(\varepsilon) D\|_2 \approx \|D\|_2^2 \cdot \sqrt{\frac{d_e}{n}}$ w.h.p. We introduce the scaled diagonal matrix $\bar{D} = \frac{D}{\|D\|_2}$. Note that $\|\bar{D}\|_F^2 = d_e$ and $\|\bar{D}\|_2 = 1$.
\begin{lemma} 
\label{lemmarownorms}
Let $e_j$ be the $j$-th vector of the canonical basis in $\real^n$. Then,
\begin{align}
    \mathbb{P}\big\{ \max_{j=1,\hdots,n} \|e_j^\top H \mbox{diag}(\varepsilon) U \bar{D}\| \gre \sqrt{\frac{d_e}{n}} + \sqrt{\frac{8 \log(n/\delta)}{n}} \big\} \less \delta\,.
\end{align}
\end{lemma}  
In contrast to the analysis of J.~Tropp~\cite{tropp2011improved} which is based on a matrix Chernoff bound, we use here a matrix Bernstein bound. To our knowledge, \cref{theorembernstein} is novel. It is a close adaptation of the matrix Bernstein bound in~\cite{tropp2015introduction} (see Theorem~6.6.1 therein) to the sub-sampled case. 
\begin{theorem}[Matrix Bernstein] 
\label{theorembernstein}
Let $\mathcal{X} = \{X_1, \hdots, X_n\}$ be a finite set of squared matrices with dimension $d$. Fix a dimension $m$, and suppose that there exists a positive semi-definite matrix $V$ and a real number $K > 0$ such that $\mathbb{E}\{X_I\} = 0$, $\mathbb{E}\{X_I^2\} \preceq V$, and $\|X_I\|_2 \less K$ almost surely, where $I$ is a uniformly random index over $\{1,\hdots,n\}$. Let $T$ be a subset of $\{1,\hdots,n\}$ with $m$ indices drawn uniformly at random without replacement. Then, for any $t \gre \sqrt{m \|V\|_2} + K/3$, we have
\begin{align*}
    \mathbb{P} \Big\{ \big\| \sum_{i \in T} X_i \big\|_2 \gre t \Big\} \, \less 8\cdot d_e \cdot \exp \big( -\frac{t^2/2}{m \|V\|_2 + K t /3} \big)\,,
\end{align*}
where $d_e \defn \mbox{tr}(V) / \|V\|$ is the intrinsic dimension of the matrix $V$.
\end{theorem} 
Equipped with the results of \cref{lemmarownorms} and \cref{theorembernstein}, we are ready to prove \cref{theoremconcentrationshrt}. We write $v_j \defn \sqrt{\frac{n}{m}}\,w_j$, where $w_j^\top = e_j^\top H \text{diag}(\varepsilon) U \bar{D}$, and $\varepsilon \in \{\pm 1\}^n$ is a fixed vector. We denote $\gamma \defn \max\big\{\max_{1\less j\less n} \|v_j\|, \, m^{-\frac{1}{2}} \big\}$ and $X_i \defn v_i v_i^\top - \frac{1}{m} \bar{D}^2$. Let $I$ be a uniformly random index over $\{1,\hdots,n\}$. We have 
\begin{align*}
    \mathbb{E}\{X_I\} &= \frac{n}{m} \big(\frac{1}{n}\sum_{i=1}^n \bar{D}U^\top \mbox{diag}(\varepsilon) H e_i e_i^\top H \mbox{diag}(\varepsilon) U \bar{D}\big) - \frac{1}{m} \bar{D}^2\\
    &= \frac{1}{m} \bar{D}U^\top \mbox{diag}(\varepsilon) H \underbrace{\sum_{i=1}^n e_i e_i^\top}_{= I} H \mbox{diag}(\varepsilon) U \bar{D} - \frac{1}{m} \bar{D}^2 = 0\,.
\end{align*}
The last equality holds due to the fact that $H^2 = I$, $\text{diag}(\varepsilon)^2 = I$ and $U^\top U = I$. Further, $\|v_I\|^2 \less \gamma^2$ a.s., so that $\|v_I\|^2 v_I v_I^\top \preceq \gamma^2 v_I v_I^\top$ a.s., and consequently, $\mathbb{E}\{\|v_I\|^2 v_I v_I^\top \} \preceq \frac{\gamma^2}{m} \cdot \bar{D}^2$. Thus, $\mathbb{E}\{X_I^2\} = \mathbb{E}\{\|v_I\|^2 v_I v_I^\top\} - \frac{2}{m} \bar{D}^4 + \frac{1}{m^2} \bar{D}^4 \preceq \frac{\gamma^2}{m} \bar{D}^2$. Furthermore, we have $\|X_I\| = \|v_I v_I^\top - \frac{1}{m} \bar{D}^2\| \less \max\left\{ \max_{j=1,\hdots,n} \|v_j\|^2, m^{-1}  \right\} = \gamma^2$.

Let $T$ be a subset of $m$ indices in $\{1, \hdots, n\}$ drawn uniformly at random, without replacement. Applying \cref{theorembernstein} with $V = m^{-1} \gamma^2 \bar{D}^2$ and using the scale invariance of the effective dimension, we obtain that $\mathbb{P} \left\{ \Big\| \sum_{i \in T} X_i \Big\|_2 \gre t \right\} \,\less 8 d_e \cdot \exp \left( -\frac{t^2/2}{\gamma^2 (1 + t /3)} \right)$ for any $t \gre \gamma + \gamma^2/3$,

Suppose now that $\varepsilon$ is a vector of independent Rademacher variables. Note that $\sum_{i \in T} X_i \overset{\mathrm{d}}{=} \bar{D} U^\top (S^\top S - I) U \bar{D}$. From \cref{lemmarownorms}, we know that $\gamma \less \sigma \defn \sqrt{\frac{d_e}{m}} + \sqrt{\frac{8 \log(2n/\delta)}{m}}$ with probability at least $1-\delta/2$. Consequently, with probability at least $1 - \delta/2 - 8d_e \cdot \exp \left( -\frac{t^2/2}{\sigma^2 (1 + t /3)} \right)$, we have $\| \bar{D} U^\top (S^\top S - I) U \bar{D}\|_2 \less t$, for $t \gre \sigma \,(1+\sigma/3)$. We set $t=\max\{4 \sigma \sqrt{\log(16d_e / \delta)}, 16\sigma^2 \log(16d_e/\delta)\}$. We have $t \gre \frac{4}{3} \max\{\sigma, \sigma^2\}$ and $\frac{t^2/2}{\sigma^2 (1+t/3)} \gre \log(16d_e /\delta)$. Therefore, $\|D U^\top (S^\top S - I) U D\|_2 \less t$ with probability at least $1-\delta/2-8 d_e \exp(-\log(16d_e/\delta)) = 1-\delta$. Equivalently, setting $t = \max\{\sqrt{m_\delta/m}, m_\delta/m\}$ where 
\begin{align*}
    \sqrt{m_\delta} = 4(\sqrt{d_e} + \sqrt{8\log(2n/\delta)}) \sqrt{\log(16d_e/\delta)}\,, 
\end{align*}
we have with probability at least $1-\delta$ that $\|C_S - I_d\|_2 \less \max\left\{\sqrt{\frac{m_\delta}{m}}, \frac{m_\delta}{m}\right\}$ and this concludes the proof of \cref{theoremconcentrationshrt}.

\subsubsection{Proof of \cref{lemmarownorms}}

We fix a row index $j \in \{1,\hdots,n\}$, and define the function $f(x) \defn \|e_j^\top H \text{diag}(x) U \bar{D}\| = \|x^\top E U \bar{D}\|$ where $E \defn \text{diag}(e_j^\top H)$. Each entry of $E$ has magnitude $n^{-\frac{1}{2}}$. The function $f$ is convex, and its Lipschitz constant is upper bounded as follows,
\begin{align*}
    |f(x)-f(y)| \less \|(x-y)^\top E U \bar{D}\| \less \|x-y\|\, \|E\|_2\, \|U\|_2\, \|\bar{D}\|_2 = \frac{1}{\sqrt{n}} \|x-y\|\,.
\end{align*}
For a Rademacher vector $\varepsilon$, we have 
\begin{align*}
    \mathbb{E}\{f(\varepsilon)\} \less \sqrt{\mathbb{E} \{f(\varepsilon)^2\}} = \|EU\bar{D}\|_F \less \|EU\|_2 \, \|\bar{D}\|_F =  \sqrt{\frac{d_e}{n}}\,.
\end{align*}
Applying Lipschitz concentration results for Rademacher variables, we obtain
\begin{align*}
    \mathbb{P}\Big\{\|e_j^\top H \text{diag}(\varepsilon) U \bar{D}\| \gre \sqrt{\frac{d_e}{n}} + \sqrt{\frac{8 \log(n/\delta)}{n}} \Big\} \less \frac{\delta}{n}\,.
\end{align*}
Finally, taking a union bound over $j \in \{1,\hdots,n\}$, we obtain the claimed result.

\subsubsection{Proof of \cref{theorembernstein}}

We denote $S_T \defn \sum_{i \in T} X_i$. Fix $\theta > 0$, define $\psi(t)=e^{\theta t} - \theta t - 1$, and use the Laplace matrix transform method (e.g., Proposition~7.4.1 in~\cite{tropp2015introduction}) to obtain
\begin{align*}
    \mathbb{P} \big\{ \lambda_\text{max}(S_T) \gre t \big\} \less \psi(t)^{-1} \, \mathbb{E} \, \mbox{tr} \,\psi(S_T) = (e^{\theta t} - \theta t - 1)^{-1} \mathbb{E} \, \mbox{tr}\big(e^{\theta S_T} - I\big)\,,
\end{align*}
and the last equality holds due to the fact that $\mathbb{E}\{S_T\} = m \,\mathbb{E}\{X_I\} = 0$. Let $T^\prime = \{i_1, \hdots, i_m\}$ be a subset of $\{1, \hdots, n\}$, drawn uniformly at random with replacement. In particular, the indices of $T^\prime$ are independent random variables, and so are the matrices $\{X_{i_j}\}_{j=1}^m$. Write $S_{T^\prime} \defn \sum_{j=1}^m X_{i_j}$. Gross and Nesme~\cite{gross2010note} have shown that $\mathbb{E}\, \mbox{tr} \exp\big( \theta S_T \big) \, \less \, \mathbb{E}\, \mbox{tr} \exp \big( \theta S_{T^\prime} \big)$ for any $\theta > 0$. As a consequence of Lieb's inequality (e.g., Lemma 3.4 in \cite{tropp2015introduction}), it holds that
\begin{align*}
    \mathbb{E}\, \mbox{tr} \exp \big( \theta S_{T^\prime} \big) \, \less \mbox{tr} \exp \big( \sum_{j=1}^m \log \, \mathbb{E}\, e^{\theta X_{i_j}} \big) = \mbox{tr} \exp \big( m \, \log \, \mathbb{E}\, e^{\theta X_I} \big)\,.
\end{align*}
Thus, it remains to bound $\mathbb{E}\{e^{\theta X_I}\}$. By assumption, $\mathbb{E}\{X_I\} = 0$ and $\|X_I\|_2 \less K$ almost surely. Then, using Lemma 5.4.10 from \cite{vershynin2018high}, we get $\mathbb{E}\{ e^{\theta X_I}\} \preceq \exp\!\left(g(\theta) \, \mathbb{E}\{X_I^2\} \right)$, for any $|\theta| < 3/K$ and where $g(\theta) = \frac{\theta^2/2}{1-|\theta| K /3}$. By monotonicity of the logarithm, we have $m \cdot \log \mathbb{E}\,e^{\theta X_I} \preceq m \cdot g(\theta) \, \mathbb{E}\{X_I^2\}$. By assumption, $\mathbb{E}\{X_I^2\} \preceq V$ and thus, $m \cdot \log \mathbb{E}\{e^{\theta X_I}\}\preceq m \cdot g(\theta) \, V$. By monotonicity of the trace exponential, it follows that 
\begin{align*}
    \mbox{tr} \exp \big( m \, \log \, \mathbb{E}\{e^{\theta X_I}\} \big) \less \mbox{tr} \exp \big(m \,g(\theta)\, V\big)\,, 
\end{align*}
and this further implies that
\begin{align*}
    \mathbb{P} \big\{ \lambda_\text{max}(S_T) \gre t \big\}  \less  \frac{1}{e^{\theta t} - \theta t - 1} \mbox{tr} \big( e^{m \,g(\theta)\, V} - I \big) = \frac{1}{e^{\theta t} - \theta t - 1} \,\mbox{tr} \, \varphi(m\,g(\theta)\,V) \,,
\end{align*}
where $\varphi(a) = e^a - 1$. The function $\varphi$ is convex, and the matrix $m \,g(\theta)\, V$ is positive semidefinite. Therefore, we can apply Lemma 7.5.1 from \cite{tropp2015introduction} and obtain 
\begin{align*}
    \mbox{tr} \, \varphi(m\,g(\theta)\,V) \less d_e \cdot \varphi(m\,g(\theta)\,\|V\|_2) \less d_e \cdot e^{m\,g(\theta)\,\|V\|_2}\,,
\end{align*}
which further implies that
\begin{align*}
    \mathbb{P} \left\{ \lambda_\text{max}(S_T) \gre t \right\} &\less  d_e \cdot \frac{e^{\theta t}}{e^{\theta t} - \theta t - 1} \cdot e^{-\theta t + m \,g(\theta) \cdot \|V\|_2}\\
    &\less d_e \cdot \big(1+\frac{3}{\theta^2 t^2}\big) \cdot e^{-\theta t + m\,g(\theta) \cdot \|V\|_2}\,.
\end{align*}
For the last inequality, we used the fact that $\frac{e^a}{e^a - a - 1} = 1 + \frac{1+a}{e^a - a -1} \less 1 + \frac{3}{a^2}$ for all $a \gre 0$. Picking $\theta = t / (m\,\|V\|_2 + Kt/3)$, we obtain 
\begin{align*}
    \mathbb{P} \big\{ \lambda_\text{max}(S_T) \gre t \big\} \, \less d_e \cdot \big(1+ 3\cdot \frac{(m \|V\|_2 + Kt/3)^2}{t^4}\big) \cdot \exp \big( -\frac{t^2/2}{m \|V\|_2 + K t /3} \big)\,.
\end{align*}
Under the assumption $t \gre \sqrt{m \|V\|_2} + K / 3$, the parenthesis in the above right-hand side is bounded by four, which results in 
\begin{align*}
    \mathbb{P} \big\{ \lambda_\text{max}(S_T) \gre t \big\} \, \less 4\cdot d_e \cdot \exp \big( -\frac{t^2/2}{m \|V\|_2 + K t /3} \big)\,.
\end{align*}
Repeating the argument for $-S_T$ and combining the two bounds, we obtain the claimed result.

\subsection{Proof of \cref{theoremgaussianconcentration}}
\label{sectiongaussianconcentration}

Let $G \in \real^{m \times d}$ be a matrix with i.i.d.~Gaussian entries $\mathcal{N}(0,1)$. We introduce the Gaussian processes $X_{u,v}$ and $Y_{u,v}$ indexed over $\mathcal{C} \times \real^m$:
\begin{align*}
    X_{u,v} = v^\top G u\,,\qquad Y_{u,v} = \|u\|_2 \cdot g^\top v + \|v\|_2 \cdot h^\top u\,,
\end{align*}
where $g \sim \mathcal{N}(0,I_m)$ and $h \sim \mathcal{N}(0,I_d)$ are random Gaussian vectors independent of each other and of the matrix $G$. 

\subsubsection*{Restricted maximum singular value} The proof is essentially based on the following Gaussian comparison inequality which is a consequence of the classical Slepian's inequality~\cite{ledoux2013probability}.
\begin{lemma}
\label{lemmaslepian}
Let $\psi : \mathcal{C} \times \real^m \to \real$ be a given function, and $\tau \in \real$. Then, we have
\begin{align}
\label{eqnuppertailcomparison}
    \mathbb{P}\big(\sup_{(u,v) \in \mathcal{C} \times \real^m} X_{u,v} + \psi(u,v) \gre \tau \big) \less 2 \cdot \mathbb{P}\big(\sup_{(u,v) \in \mathcal{C} \times \real^m} Y_{u,v} + \psi(u,v) \gre \tau \big)\,.
\end{align}
\end{lemma} 
We introduce the function $\psi(u,v) = -\frac{m}{2}\|u\|_2^2 - \frac{1}{2} \|v\|_2^2$. Note that, for a Gaussian embedding $\Pi \in \real^{m\times d}$ with i.i.d.~entries $\mathcal{N}(0,1/m)$, we have
\begin{align*}
    \sup_{u \in \mathcal{C}}\, \langle u, (\Pi^\top \Pi - I_d) u \rangle \overset{\mathrm{d}}{=} \frac{2}{m} \cdot \sup_{(u,v) \in \mathcal{C} \times \real^m} X_{u,v} + \psi(u,v)\,.
\end{align*}
According to \cref{lemmaslepian} and up to a factor $2/m$, the maximum eigenvalue of the matrix $(\Pi^\top \Pi - I_d)$ restricted to the set $\mathcal{C}$ is stochastically dominated by the random variable $Y \defn \sup_{(u,v)\in \mathcal{C} \times \real^m} Y_{u,v} + \psi(u,v)$. The rest of the proof amounts to characterize the upper deviations of $Y$ through Gordon-type optimization:
\begin{align*}
    Y &= \sup_{(u, v) \in \mathcal{C} \times \real^m} \|u\|_2 \cdot g^\top v + \|v\|_2 \cdot h^\top u - \frac{m}{2} \|u\|_2^2 - \frac{1}{2}\|v\|_2^2\\
    &= \sup_{u \in \mathcal{C},\,t \gre 0} t \cdot (\|u\|_2 \|g\|_2 + h^\top u) - \frac{t^2}{2} - \frac{m}{2} \|u\|_2^2\\
    &= \sup_{u \in \mathcal{C}} \, \frac{1}{2} (\|u\|_2 \|g\|_2 + h^\top u)_+^2 - \frac{m}{2} \|u\|_2^2\\
    &\less \sup_{u \in \mathcal{C}} \, \frac{1}{2} (\|u\|_2 \|g\|_2 + |h^\top u|)^2 - \frac{m}{2} \|u\|_2^2\\
    &= \sup_{u \in \mathcal{C}} \, \frac{1}{2} \big(\|u\|_2 (\|g\| - \sqrt{m}) + |h^\top u|\big) \big(\|u\|_2 (\|g\| + \sqrt{m}) + |h^\top u| \big)\,.
\end{align*}
Consider the event $\mathcal{E} = \{|\|g\|_2 - \sqrt{m}| \less t\}$. From classical Gaussian concentration bounds~\cite{ledoux2013probability, vershynin2018high}, we know that $\mathbb{P}(\mathcal{E}) \gre 1-2 e^{-t^2/2}$. Conditional on $\mathcal{E}$ and using the fact that $\text{rad}(\mathcal{C}) \less 1$, we have:
\begin{align*}
    Y \less \sup_{u \in \mathcal{C}}\, \frac{1}{2} (t + |h^\top u|) ((2\sqrt{m}+t) + |h^\top u|) = \frac{1}{2} \cdot \Big( \big(\sqrt{m}+t+\sup_{u \in \mathcal{C}} |h^\top u|\big)^2 - m \Big)\,.
\end{align*}
Denote $f(w) = \sup_{u \in \mathcal{C}} |w^\top u|$. This function is $1$-Lipschitz, whence $|f(h) - \mathbb{E}[f(h)]| \less t$ with probability $1-2e^{-t^2/2}$. This implies that
\begin{align}
\label{eqnintermediatemaxgaussian}
    \sup_{u \in \mathcal{C}} \, \langle u, (\Pi^\top \Pi - I_d) u\rangle \less \big(1+\frac{\omega(\mathcal{C}) + 2t}{\sqrt{m}}\big)^2 - 1\,,
\end{align}
with probability at least $1-8e^{-t^2/2}$.

\subsubsection*{Restricted minimum singular value} The proof essentially leverages Gordon's minimax theorem~\cite{gordon1985some}, and more precisely, the Gordon's minimax theorem in the presence of convexity~\cite{thrampoulidis2014tight}.

We introduce the random variable $Y = \inf_{u \in \mathcal{C}} \sup_{v \in \real^m} \left\{ Y_{u,v} + \psi(u,v) \right\}$, and we have
\begin{align*}
    Y &\gre \inf_{u \in \mathcal{C}} \, \frac{1}{2} (h^\top u + \|u\|_2 \|g\|_2)_+^2 - \frac{m}{2}\|u\|_2^2\\
    & \gre \inf_{u \in \mathcal{C}} \, \frac{1}{2} (-\sup_{u \in \mathcal{C}} |h^\top u| + \|u\|_2 \|g\|_2)_+^2 - \frac{m}{2}\|u\|_2^2\\
    & \gre \frac{m}{2} \cdot \inf_{u \in \mathcal{C}} \, (-\frac{\omega(\mathcal{C})+2t}{\sqrt{m}} + \|u\|_2)_+^2 - \|u\|_2^2\,.
\end{align*}
where the last inequality holds conditional on the event %
\begin{align*}
    \{|\|g\|_2 - \sqrt{m}| \less t\} \cap \{|\sup_{u \in \mathcal{C}} |h^\top u| - \omega(\mathcal{C})| \less t\}
\end{align*}
which holds with probability at least $1-4e^{-t^2/2}$. For two positive numbers $\alpha, \beta$, it always holds that $(\alpha - \beta)_+^2 - \alpha^2 \gre -2 \max\{\alpha, \beta\} \beta + \beta^2$. Setting $\alpha = \|u\|_2$ and $\beta = \frac{\omega(C)+2t}{\sqrt{m}}$ and using that $\alpha \less 1$, we obtain
\begin{align*}
    \frac{2}{m} \cdot Y \gre -2 \max\{1, \frac{\omega(\mathcal{C}) + 2t}{\sqrt{m}}\} \cdot \frac{\omega(\mathcal{C}) + 2t}{\sqrt{m}} + \left(\frac{\omega(\mathcal{C}) + 2t}{\sqrt{m}}\right)^2\,,
\end{align*}
with probability $1-4e^{-t^2/2}$. Finally, it follows from Theorem II.1 in~\cite{thrampoulidis2014tight} (the Gaussian min-max theorem in the presence of convexity) that for any $\tau \in \real$,
\begin{align*}
    \mathbb{P}\big( \frac{2}{m} \cdot \inf_{u \in \mathcal{C}} \sup_{v \in \real^m}\, v^\top G u + \psi(u,v) < \tau \big) \less 2\cdot \mathbb{P}\big( \frac{2}{m} \cdot Y < \tau \big)
\end{align*}
We note that $\frac{2}{m} \cdot \inf_{u \in \mathcal{C}} \sup_{v \in \real^m}\, v^\top G u + \psi(u,v) \overset{\mathrm{d}}{=} \inf_{u \in \mathcal{C}} \langle u, (\Pi^\top \Pi - I_d) u\rangle$, and this yields the claimed inequality
\begin{align}
\label{eqnintermediatemingaussian}
    \inf_{u \in \mathcal{C}} \langle u, (\Pi^\top \Pi - I_d) u\rangle \gre -2 \max\{1, \frac{\omega(\mathcal{C}) + 2t}{\sqrt{m}}\} \cdot \frac{\omega(\mathcal{C}) + 2t}{\sqrt{m}} + \big(\frac{\omega(\mathcal{C}) + 2t}{\sqrt{m}}\big)^2\,,
\end{align}
with probability at least $1-8e^{-t^2/2}$. Using a union bound over the deviation bounds \eqref{eqnintermediatemaxgaussian} and \eqref{eqnintermediatemingaussian} and setting $t = \sqrt{2 \log(16/\delta)}$, we obtain the claim.

\subsubsection{Proof of \cref{lemmaslepian}}

We introduce the auxiliary Gaussian process $X^\prime_{u,v} = v^\top G u + Z \|u\|_2 \|v\|_2$, where $Z \sim \mathcal{N}(0,1)$ is independent of $g,h,G$.
Note that the Gaussian processes $X^\prime_{u,v}$ and $Y_{u,v}$ have zero-mean, i.e., $\mathbb{E}[X^\prime_{u,v}] = \mathbb{E}[Y_{u,v}] = 0$ for all $(u,v) \in \mathcal{C} \times \real^m$, and that their second moments are equal, i.e.,
\begin{align*}
    \mathbb{E}[{X^\prime_{u,v}}^2]= 2 \|u\|_2^2 \|v\|_2^2 = \mathbb{E}[Y_{u,v}^2]\,.
\end{align*}
Fix $(u, v), (u^\prime, v^\prime) \in \mathcal{C} \times \real^m$. We have
\begin{align*}
    &\mathbb{E}[X^\prime_{u,v} X^\prime_{u^\prime,v^\prime}] = \|u\|_2\|u^\prime\|_2\|v\|_2\|v^\prime\|_2 + u^\top u^\prime v^\top v^\prime\\
    & \mathbb{E}[Y_{u,v} Y_{u^\prime,v^\prime}] = \|u\|_2 \|u^\prime\|_2 v^\top v^\prime + \|v\|_2 \|v^\prime\|_2 u^\top u^\prime\,.
\end{align*}
Consequently, we have
\begin{align*}
    \mathbb{E}[X^\prime_{u,v} X^\prime_{u^\prime, v^\prime}] - \mathbb{E}[Y_{u,v} Y_{u^\prime, v^\prime}] = (\|u\|_2 \|u^\prime\|_2 - u^\top u^\prime) (\|v\|_2 \|v^\prime\|_2 - v^\top v^\prime) \gre 0\,.
\end{align*}
From Slepian's inequality (see Ch.~3 in~\cite{ledoux2013probability}), we have for any $\tau \in \real$ that
\begin{align*}
    \mathbb{P}\big(\sup_{(u,v) \in \mathcal{C} \times \real^m} X^\prime_{u,v} + \psi(u,v) \gre \tau \big) \less \mathbb{P}\big(\sup_{(u,v) \in \mathcal{C} \times \real^m} Y_{u,v} + \psi(u,v) \gre \tau \big)\,.
\end{align*}
On the other hand, it holds that
\begin{align*}
    \mathbb{P}\big( \{\sup_{(u,v) \in \mathcal{C} \times \real^m} v^\top Gu + \psi(u,v) \gre \tau\} \cap \{Z > 0\} \big) \less \mathbb{P}\big(\sup_{(u,v) \in \mathcal{C} \times \real^m} X^\prime_{u,v} + \psi(u,v)\gre \tau \big)
\end{align*}
The events $\{\sup_{(u,v) \in \mathcal{C} \times \real^m} v^\top Gu + \psi(u,v) \gre \tau\}$ and $\{Z > 0\}$ are independent, whence 
\begin{small}
\begin{align*}
    \mathbb{P}\big( \{\sup_{(u,v) \in \mathcal{C} \times \real^m} v^\top Gu + \psi(u,v) \gre \tau\} \cap \{Z > 0\} \big) = \frac{1}{2} \cdot \mathbb{P}\big(\sup_{(u,v) \in \mathcal{C} \times \real^m} v^\top Gu + \psi(u,v) \gre \tau \big)\,. 
\end{align*}
\end{small}
Combining the previous inequalities, we obtain the claimed result, that is,
\begin{align*}
    \mathbb{P}\big(\sup_{(u,v) \in \mathcal{C} \times \real^m} X_{u,v} + \psi(u,v) \gre \tau \big) \less 2 \cdot \mathbb{P}\big(\sup_{(u,v) \in \mathcal{C} \times \real^m} Y_{u,v} + \psi(u,v) \gre \tau \big)\,.
\end{align*}

\section*{Acknowledgments}
This work was partially supported by the National Science Foundation under grants IIS-1838179 and ECCS-2037304, Facebook Research, Adobe Research and Stanford SystemX Alliance.

\bibliographystyle{siamplain}

\appendix

\section{Heavy-ball Acceleration and Adaptivity}
\label{sectionchebychev}

It is of interest to consider whether the pre-conditioned heavy-ball update
\begin{align}
	\label{eqnpolyakihs}
	x_{t+1} = x_t - \mu H_S^{-1} \nabla f(x_t) + \beta (x_t - x_{t-1})
\end{align}
could be made adaptive. The update~\eqref{eqnpolyakihs} is also known in the literature as the second-order Richardson method or the heavy-ball method with preconditioner $H_S$. We refer to it as the Polyak IHS update. Although we will see that this is theoretically possible, it leads to a non-practical algorithm. Nonetheless, we believe that our analysis in this section is of independent interest, and may be refined to obtain a numerically efficient adaptive version of the Polyak-IHS method.

\subsection{Non-asymptotic Guarantees for the Heavy-ball Method}

Standard analyses~\cite{polyak1964some} of the heavy-ball method applied to a quadratic objective function typically give \textit{asymptotic guarantees} (or local guarantees) of the form 
\begin{align}
	\lim_{t \to \infty} (\delta_{t} / \delta_0)^{1/t} \less (\frac{\sqrt{\kappa} - 1}{\sqrt{\kappa}+1})^2\,,
\end{align}
where $\kappa$ is the condition number of the quadratic objective function, for an appropriate choice of the step size $\mu$ and momentum parameter $\beta$. On the other hand, non-asymptotic (but non-explicit or weaker) guarantees exist in the literature.~\cite{lessard2016analysis} established (see their Proposition~2) the non-asymptotic rate $\delta_t \less \kappa_P \cdot (\frac{\sqrt{\kappa} - 1}{\sqrt{\kappa}+1})^{2t} \cdot \delta_0$ where $\kappa_P$ is the condition number of a positive definite matrix $P \in \real^{d \times d}$ solution of a linear matrix inequality (LMI) that, in our case, would take $\mathcal{O}(nd^2)$ flops to form and solve, and this defeats the purpose of sketching.~\cite{ghadimi2015global} established that for a $\mu$-strongly convex objective function with $L$-Lipschitz continuous gradient (but not necessarily twice differentiable), the heavy-ball method has linear convergence rate $q$, but with $q > (\frac{\sqrt{\kappa} - 1}{\sqrt{\kappa}+1})^2$. Non-asymptotic linear convergence guarantees have also been provided for the stochastic heavy-ball method (see Theorem~1 in~\cite{loizou2020momentum}), but the linear rate is also greater than the accelerated rate $(\frac{\sqrt{\kappa} - 1}{\sqrt{\kappa}+1})^2$.

We provide here a novel contribution, by expliciting a near-linear convergence rate for the heavy-ball method which is asymptotically linear and equal to the accelerated rate $(\frac{\sqrt{\kappa} - 1}{\sqrt{\kappa}+1})^2$. Our technical innovation is to leverage Theorem~1 in~\cite{kozyakin2009accuracy}, which characterizes the approximation error $\rho(M) \approx \|M^t\|_2^{1/t}$, where $\rho(M)$ is the spectral radius of a matrix $M$. 
\begin{lemma}
	\label{lemmapolyakdynamics}
	Consider a dynamical system of the form
	\begin{align}
		\label{eqnpolyakdynamics}
		\begin{bmatrix} z_{t+1}\\ z_t \end{bmatrix} =  X_{\mu,\beta} \begin{bmatrix} z_t \\ z_{t-1} \end{bmatrix} \quad \mbox{where} \quad X_{\mu, \beta} \defn \begin{bmatrix} (1+\beta)I_d - \mu X & -\beta I_d \\ I_d & 0 \end{bmatrix}\,,
	\end{align}
	where $X$ is a positive definite matrix with $\lambda \cdot I_d \preceq X \preceq \Lambda \cdot I_d$, and $\mu = \frac{4}{(\sqrt{\lambda} + \sqrt{\Lambda})^2}$ and $\beta = \left(\frac{\sqrt{\Lambda} - \sqrt{\lambda}}{\sqrt{\Lambda} + \sqrt{\lambda}}\right)^2$. Then, it holds for any $t \gre 1$ that
	\begin{align}
		\frac{\|z_{t+1}\|_2^2 + \|z_t\|_2^2}{\|z_1\|_2^2 + \|z_0\|_2^2} \less \alpha(t) \cdot \beta^{\omega(t)}\,, 
	\end{align}
	where $\alpha(t) = 3^{\nu(t) (\nu(t)+1)} \cdot (1+4\beta + \beta^2)^{2\nu(t)}$ and $\omega(t) = t-2\nu(t)$, and $\nu(t) = \log(t)/\log(2) + 1$. Consequently, we obtain the asymptotic error
	\begin{align}
		\lim_{t \to +\infty} \, \left(\frac{\|z_t\|_2^2}{\|z_0\|_2^2}\right)^{1/t} \less \left(\frac{\sqrt{\Lambda} - \sqrt{\lambda}}{\sqrt{\Lambda} + \sqrt{\lambda}}\right)^2\,.
	\end{align}
\end{lemma} 

\subsection{Non-asymptotic Guarantees for Polyak-IHS}

Multiplying both sides of the update~\eqref{eqnpolyakihs} by $H^\frac{1}{2}$ and subtracting $H^\frac{1}{2}x^*$, we obtain the error recursion $\Delta_{t+1} = (I_d - \mu C_S^{-1}) \Delta_t + \beta (\Delta_t - \Delta_{t-1})$. We have in the equivalent matrix form
\begin{align}
	\label{eqnpolyakihsdynamics}
	\begin{bmatrix} \Delta_{t+1}\\ \Delta_t \end{bmatrix} =  M_{\mu,\beta} \begin{bmatrix} \Delta_t \\ \Delta_{t-1} \end{bmatrix}\quad \mbox{where} \quad M_{\mu, \beta} \defn \begin{bmatrix} (1+\beta)I_d - \mu C_S^{-1} & -\beta I_d \\ I_d & 0 \end{bmatrix}\,,
\end{align}
Combining~\eqref{eqnpolyakihsdynamics} and Lemma~\ref{lemmapolyakdynamics}, we obtain the following non-asymptotic guarantee.
\begin{corollary}
	\label{corollaryfinitetimeguaranteespolyak}
	Consider the step size $\mu_\rho = \frac{2(1-\rho)}{1+\sqrt{1-\rho}}$ and momentum parameter $\beta_\rho = \frac{1-\sqrt{1-\rho}}{1+\sqrt{1-\rho}}$. Then, conditional on $\mathcal{E}_\rho$, the Polyak-IHS update~\eqref{eqnpolyakihs} satisfies for any $t \gre 1$ the finite-time guarantee 
	\begin{align}
		\label{eqnfinitetimeguaranteespolyak}
		\boxed{\frac{\delta_{t+1} + \delta_t}{\delta_1 + \delta_0} \less \alpha(t,\rho) \cdot \beta_\rho^{\omega(t)}}\,,    
	\end{align}
	where $\alpha(t,\rho) = 3^{\nu(t) (\nu(t)+1)} \cdot (1+4\beta_\rho + \beta_\rho^2)^{2\nu(t)}$, $\omega(t) = t-2\nu(t)$, and $\nu(t) = \log(t)/\log(2) + 1$.
\end{corollary} 
As an immediate consequence of Corollary~\ref{corollaryfinitetimeguaranteespolyak}, we recover the standard asymptotic guarantee
\begin{align}
	\label{eqnasymptoticguaranteespolyak}
	\limsup_{t \to +\infty} \left(\frac{\delta_{t}}{\delta_0}\right)^{\frac{1}{t}} \less \beta_\rho = \frac{\rho}{4} \cdot (1+o(1))\,.
\end{align}
The Polyak-IHS update yields asymptotically the same accelerated rate as PCG. However, the finite-time guarantee~\eqref{eqnfinitetimeguaranteespolyak} becomes significant after a large number of iterations (see Table~\ref{tablecoefficientspectralradiusapproximation}). For instance, in order to guarantee finite-time convergence faster than the IHS, i.e., $\alpha(t,\rho) \beta_\rho^{\omega(t)} \less \rho^t$, then at least $t \gtrsim 100$ iterations are needed for $\rho \in \{0.1, 0.05, 0.01, 0.001\}$. In other words, testing the hypothesis that the sketch size is large enough to guarantee convergence faster than the IHS would require at least $t \gtrsim 100$ iterations for these values of $\rho$. This is impractical for many settings of interest: for $\rho \lesssim 0.1$, least-squares solvers reach double precision within $\lesssim 20$ iterations).
\begin{table}[!h]
	\caption{Upper bound $(\alpha(t,\rho) \beta_\rho^{\omega(t)})^{1/t}$ for different values of $\rho$ and $t$. Bold values correspond to convergence guaranteed to be faster than the IHS update, i.e., $\alpha(t,\rho) \beta_\rho^{\omega(t)} \less \rho^t$.}
	\label{tablecoefficientspectralradiusapproximation}
	\centering
	\begin{scriptsize}
		\begin{tabular}{|c|c|c|c|c|c|c|c|}
			\cmidrule(r){1-8}
			& $t=1$ & $t=10$ & $t=50$ & $t=100$ & $t=200$ & $t=300$ & $t=\infty$\\
			\midrule
			$\rho=0.1$ & $4.2 \cdot 10^2$ & $8.3$ & $2.1 \cdot 10^{-1}$ & $\mathbf{9.6 \cdot 10^{-2}}$ & $\mathbf{5.7 \cdot 10^{-2}}$ & $\mathbf{4.7 \cdot 10^{-2}}$ & $\mathbf{2.6 \cdot 10^{-2}}$ \\
			$\rho=0.05$ & $7.75 \cdot 10^2$ & $7.2$ & $1.2 \cdot 10^{-1}$ & $5.2 \cdot 10^{-2}$ & $\mathbf{3.0 \cdot 10^{-2}}$ & $\mathbf{2.3 \cdot 10^{-2}}$ & $\mathbf{1.2 \cdot 10^{-2}}$ \\
			$\rho=0.01$ & $3.6 \cdot 10^4$ & $5.6$ & $3.7 \cdot 10^{-1}$ & $1.3 \cdot 10^{-2}$ & $\mathbf{6.7 \cdot 10^{-3}}$ & $\mathbf{5.1 \cdot 10^{-3}}$ & $\mathbf{2.5\cdot 10^{-3}}$\\
			$\rho=0.001$ & $3.6 \cdot 10^4$ & $4.1$ & $6.9 \cdot 10^{-3}$ & $1.8 \cdot 10^{-3}$ & $\mathbf{8 \cdot 10^{-4}}$ & $\mathbf{5.9 \cdot 10^{-4}}$ & $\mathbf{2. \cdot 10^{-4}}$\\
			\bottomrule
		\end{tabular}
	\end{scriptsize}
\end{table}

\subsection{Proofs of technical results in Section~\ref{sectionchebychev}}

\subsubsection*{Proof of Lemma~\ref{lemmapolyakdynamics}}

From~\eqref{eqnpolyakdynamics}, we immediately find the recursion $\frac{\|z_{t+1}\|_2^2 + \|z_{t}\|_2^2}{\|z_{1}\|_2^2 + \|z_{0}\|_2^2} \less \|X_{\mu,\beta}^t\|_2^2$. Let $X = T \Lambda T^\top$ be an eigenvalue decomposition, where $\Lambda = \textrm{diag}(\gamma_1, \dots, \gamma_d)$ and $\gamma_1 \gre \dots \gamma_d > 0$, and define the $(2d) \times (2d)$ permutation matrix $\Pi$ as $\Pi_{i,j} = 1$ if $i$ is odd and $j=i$; $\Pi_{i,j} = 1$ if $i$ is even and $j=d+i$ and $\Pi_{i,j}=0$ otherwise. Then, it holds that
\begin{align*}
	\Pi \begin{bmatrix} T & 0 \\ 0 & T \end{bmatrix}^\top X_{\mu,\beta} \begin{bmatrix} T & 0 \\ 0 & T \end{bmatrix} \Pi^\top = \begin{bmatrix} X^{(1)}_{\mu,\beta} & 0 & \dots & 0 \\ 0 & X^{(2)}_{\mu,\beta} & \dots & 0 \\ \vdots & & \ddots & \vdots \\ 0 & 0 & \dots & X^{(d)}_{\mu,\beta} \end{bmatrix}\,,
\end{align*}
where $X^{(i)}_{\mu,\beta} = \begin{bmatrix} 1+\beta - \mu \gamma_i & -\beta \\ 1 & 0 \end{bmatrix}$. That is, $X_{\mu,\beta}$ is similar to the block diagonal matrix with $2 \times 2$ diagonal blocks $X^{(i)}_{\mu,\beta}$. For $i \in [d]$, the eigenvalues of the $2 \times 2$ matrix are roots of the equation $u^2 - (1+\beta - \mu \gamma_i)u + \beta = 0$. In the case that $1 \gre \beta \gre (1-\sqrt{\mu \gamma_i})^2$, the roots of the characteristics equations are imaginary, and both have magnitude $\sqrt{\beta}$. Then, we have that $\beta \gre (1-\sqrt{\mu \gamma_i})^2$ for all $i \in [d]$, so that the spectral norm of $X^{(i)}_{\mu,\beta}$ denoted by $\rho(X^{(i)}_{\mu,\beta})$ satisfies $\rho(X^{(i)}_{\mu,\beta}) \less \sqrt{\beta}$. On the other hand, the block-diagonal decomposition yields that 
\begin{align}
	\label{eqnintermediatepolyak1}
	\frac{\|z_{t+1}\|_2^2 + \|z_{t}\|_2^2}{\|z_{1}\|_2^2 + \|z_{0}\|_2^2} \less \max_{i \in [d]} \|(X^{(i)}_{\mu,\beta})^t\|_2^2\,.   
\end{align}
We leverage the following result, which provides an approximation error for Gelfand's formula.
\begin{theorem}[Theorem~1 in~\cite{kozyakin2009accuracy}]
	\label{theoremaccuracyspectralradius}
	Let $Z \in \real^{2 \times 2}$. It holds for any $t \gre 1$ that
	\begin{align}
		\|Z^t\|_2^2 \less 3^{\nu(t) (\nu(t)+1)} \cdot \left(\frac{\|Z\|^2_2}{\|Z^2\|_2}\right)^{2\nu(t)} \cdot \rho(Z)^{2t}\,,
	\end{align}
	where $\nu(t) = \log(t)/\log(2) + 1$.
\end{theorem} 
Using Theorem~\ref{theoremaccuracyspectralradius}, it follows from~\eqref{eqnintermediatepolyak1} that
\begin{align}
	\label{eqnpolyakfinitesampleanalysisintermediate}
	\frac{\|z_{t+1}\|_2^2 + \|z_{t}\|_2^2}{\|z_{1}\|_2^2 + \|z_{0}\|_2^2} \less 3^{\nu(t) (\nu(t)+1)} \cdot \max_{i=1,\dots, d} \left(\frac{\|X^{(i)}_{\mu,\beta}\|^2_2}{\|(X^{(i)}_{\mu,\beta})^2\|_2}\right)^{2\nu(t)} \cdot \beta^t\,.
\end{align}
We leverage the following technical result.
\begin{lemma} 
	\label{lemmaratiopolyakeigs}
	It holds that $\max_{i \in [d]} \frac{\|X^{(i)}_{\mu,\beta}\|^2_2}{\|(X^{(i)}_{\mu,\beta})^2\|_2} \less \frac{1}{\beta} \cdot (1 + 4 \beta + \beta^2)$.
\end{lemma} 
Plugging-in the upper bound of Lemma~\ref{lemmaratiopolyakeigs} into~\eqref{eqnpolyakfinitesampleanalysisintermediate}, we obtain the claimed result and this concludes the proof of Lemma~\ref{lemmapolyakdynamics}.

\subsubsection*{Proof of Lemma~\ref{lemmaratiopolyakeigs}}
For $i\in [d]$, denote $a_i = 1+\beta - \mu \gamma_i$. We have that $(X^{(i)}_{\mu,\beta})^2 = \begin{bmatrix} a_i^2 -\beta & -a_i \beta \\ a_i & -\beta \end{bmatrix}$. Therefore, $\|(X^{(i)}_{\mu,\beta})^2\|_2 \gre \beta$. Further, $\|X^{(i)}_{\mu,\beta}\|_2^2 \less \|X^{(i)}_{\mu,\beta}\|_F^2 = a_i^2 + \beta^2 + 1$. Note that $|a_i| = 2 |\frac{\lambda + \Lambda - 2 \gamma_i}{(\sqrt{\Lambda} + \sqrt{\lambda})^2}|$, whence $\max_{i\in [d]} |a_i| \less 2 \sqrt{\beta}$ and the maximum is attained for $i=1,d$. Thus, we get $\|X^{(i)}_{\mu,\beta}\|_2^2 \less 1 + \beta^2 + 4 \beta$, and this concludes the proof.

\begin{remark}
	The upper bound of Lemma~\ref{lemmaratiopolyakeigs} is tight in a worst-case sense. Note that $\|X^{(i)}_{\mu,\beta}\|_2 \gre 1$ for any $i\in [d]$. On the other hand, if there exists $\gamma_i$ such that $(\Lambda  +\lambda)/2 = \gamma_i$, then $a_i = 0$ and $\|(X^{(i)}_{\mu,\beta})^2\|_2 = \beta$. This shows that $\max_{i \in [d]} \frac{\|X^{(i)}_{\mu,\beta}\|_2^2}{\|(X^{(i)}_{\mu,\beta})^2\|_2} \gre 1/\beta$. Combining this worst-case lower bound with the upper bound of Lemma~\ref{lemmaratiopolyakeigs}, we obtain $\max_{i \in [d]} \frac{\|X^{(i)}_{\mu,\beta}\|_2^2}{\|(X^{(i)}_{\mu,\beta})^2\|_2} = \frac{1}{\beta} \cdot (1+\mathcal{O}(\beta))$.
\end{remark}

\end{document}